\pdfoutput=1

\documentclass[11pt]{article}

\usepackage[final]{acl} 

\usepackage{times}
\usepackage{latexsym}

\usepackage[T1]{fontenc}

\usepackage[utf8]{inputenc}

\usepackage{microtype}

\usepackage{inconsolata}

\usepackage{multirow}
\usepackage{booktabs}
\usepackage{amsmath}
\usepackage{CJKutf8} 
\usepackage{graphicx}
\usepackage{enumitem}
\usepackage{colortbl}
\usepackage[normalem]{ulem}
\useunder{\uline}{\ul}{}
\usepackage{dsfont}
\usepackage{hyperref}
\usepackage{textcomp}

\title{WebCiteS: Attributed Query-Focused Summarization on Chinese Web Search Results with Citations}

\author{
    Haolin Deng{\textsuperscript{$1$}}\thanks{This work was done during the internship at Tencent.}\enskip 
    Chang Wang{\textsuperscript{$2$}} \enskip 
    Xin Li{\textsuperscript{$2$}} \enskip 
    Dezhang Yuan{\textsuperscript{$2$}} \enskip 
    Junlang Zhan{\textsuperscript{$2$}} \\
    \textbf{Tianhua Zhou}{\textsuperscript{$2$}} \quad 
    \textbf{Jin Ma}{\textsuperscript{$3$}} \quad 
    \textbf{Jun Gao}{\textsuperscript{$1$}\footnotemark[2]} \quad 
    \textbf{Ruifeng Xu}\textsuperscript{$1$}\thanks{Corresponding authors} \\
    \textsuperscript{$1$}Harbin Institute of Technology, Shenzhen, China \\
    \textsuperscript{$2$}Tencent \\
    \textsuperscript{$3$}University of Science and Technology of China \\
    hldeng028@gmail.com, dezhangyuan@tencent.com, xuruifeng@hit.edu.cn
}

\begin{document}
\begin{CJK*}{UTF8}{gbsn}
\maketitle
\begin{abstract}
    Enhancing the attribution in large language models (LLMs) is a crucial task. One feasible approach is to enable LLMs to cite external sources that support their generations. However, existing datasets and evaluation methods in this domain still exhibit notable limitations.
    In this work, we formulate the task of attributed query-focused summarization~(AQFS) and present WebCiteS, a Chinese dataset featuring 7k human-annotated summaries with citations. WebCiteS derives from real-world user queries and web search results, offering a valuable resource for model training and evaluation.
    Prior works in attribution evaluation do not differentiate between groundedness errors and citation errors. They also fall short in automatically verifying sentences that draw partial support from multiple sources.
    We tackle these issues by developing detailed metrics and enabling the automatic evaluator to decompose the sentences into sub-claims for fine-grained verification.
    Our comprehensive evaluation of both open-source and proprietary models on WebCiteS highlights the challenge LLMs face in correctly citing sources, underscoring the necessity for further improvement.\footnote{The dataset and code are released under Apache License 2.0 at \url{https://github.com/HarlynDN/WebCiteS}}
    
\end{abstract}

\section{Introduction}
\label{sec:intro}
In today's information-driven society, swift access to knowledge is essential. A major limitation of web search engines is the need for users to manually compile information from various sources, which can be time-consuming.
Large language models~(LLMs)~\cite{Zhao2023ASO} exhibit potential in this domain by generating straightforward and well-organized responses. However, the potential risks of hallucination~\citep{10.1145/3571730, zhang2023sirens} and factual errors~\citep{min-etal-2023-factscore} undermine their trustworthiness as knowledge sources. 

An emerging solution is \textit{generative search engines}~\citep{Liu2023EvaluatingVI} which use LLMs to synthesize web search results into responses with in-line citations. This allows users and developers to verify the generations against the cited sources. However, recent investigations on commercial products and retrieval-augmented LLMs reveal frequent occurrences of unsupported claims and incorrect citations~\citep{Liu2023EvaluatingVI,Gao2023EnablingLL}, highlighting the challenges of attribution in LLMs~\citep{li2023survey}.
Nonetheless, the limitations of pertinent datasets and evaluation methods pose obstacles to in-depth explorations within the community.

\begin{figure}[t]
    \centering
    \small
    \includegraphics[width=0.9\linewidth]{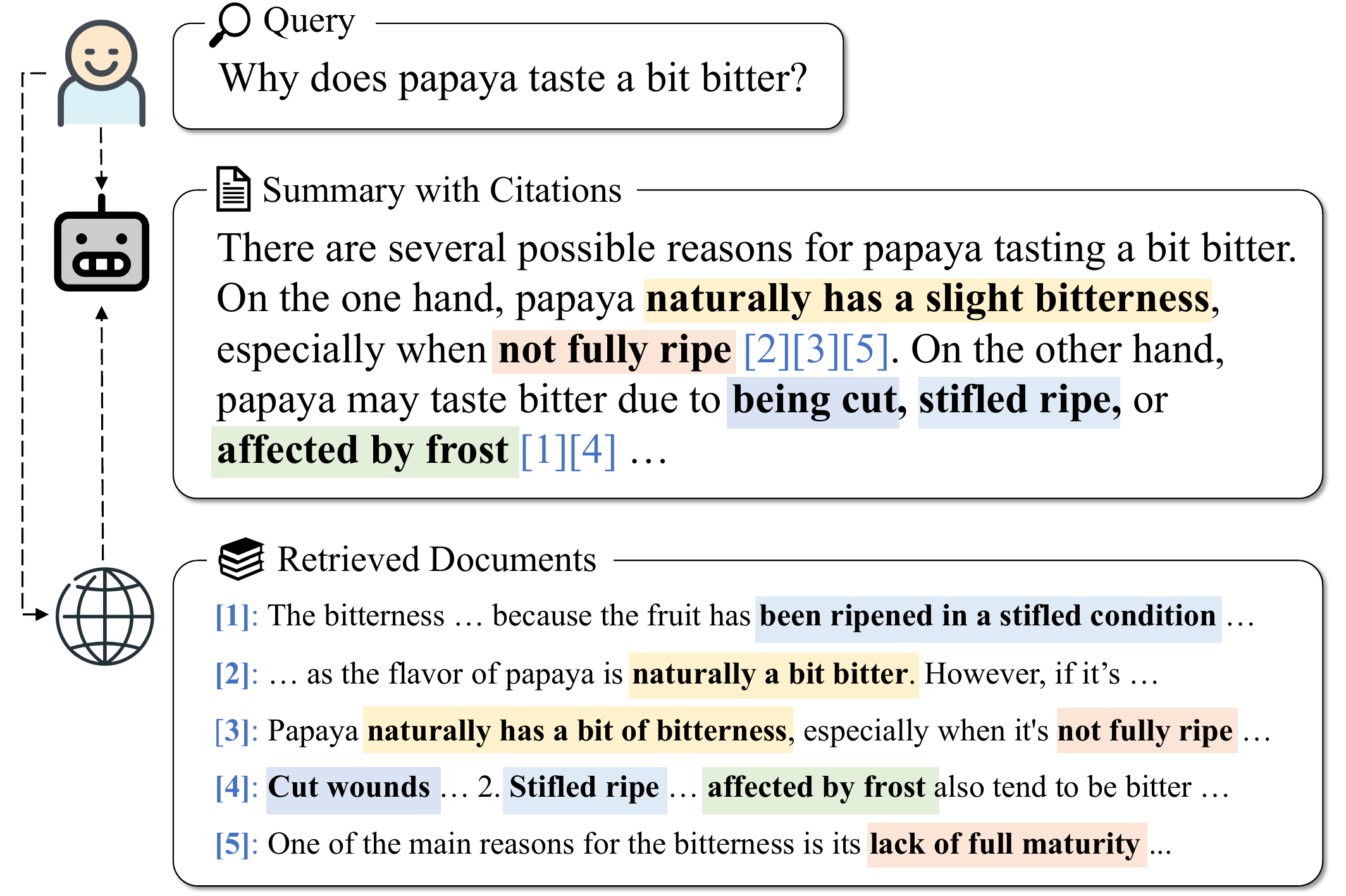}
    \caption{Illustration of attributed query-focused summarization~(AQFS). Full example is shown in Table~\ref{tab:fsp_prompt}.}
    \label{fig:intro}
\end{figure}

\begin{table*}[t]
\centering
\small
\setlength\tabcolsep{2.5pt}
\begin{tabular}{@{}cccccccccc@{}}
\toprule
 &
  \textbf{Lan.} &
  \begin{tabular}[c]{@{}c@{}}\textbf{Query}\\ \textbf{Source}\end{tabular} &
  \begin{tabular}[c]{@{}c@{}}\textbf{Document}\\ \textbf{Source}\end{tabular} &
  \textbf{\# Cnt.} &
  \begin{tabular}[c]{@{}c@{}}\textbf{Docs}\\ \textbf{Length}\end{tabular} &
  \begin{tabular}[c]{@{}c@{}}\textbf{Response}\\ \textbf{Length}\end{tabular} &
  \begin{tabular}[c]{@{}c@{}}\# \textbf{Citations}\\ \textbf{per Sent.}\end{tabular} &
  \begin{tabular}[c]{@{}c@{}}\textbf{Citation}\\ \textbf{Annot.}\end{tabular} &
  \begin{tabular}[c]{@{}c@{}}\textbf{Citation}\\ \textbf{Eval.}\end{tabular} \\ \midrule
WebCiteS &
  ZH &
  Real user queries &
  \begin{tabular}[c]{@{}c@{}}Web search\\ results\end{tabular} &
  7k &
  1025 / 3970 &
  167 &
  1.55 &
  human &
  auto \\ \midrule
\begin{tabular}[c]{@{}c@{}}WebCPM\\ \citep{qin2023webcpm}\end{tabular} &
  ZH &
  \begin{tabular}[c]{@{}c@{}}Translated \\ Reddit questions \end{tabular} &
  \begin{tabular}[c]{@{}c@{}}Web searh\\ results\end{tabular} &
  5k &
  513$^{\dagger}$ &
  244$^{\dagger}$ &
  0.34 &
  human &
  N/A \\ \midrule
\begin{tabular}[c]{@{}c@{}}WebGLM\\ \cite{liu2023webglm}\end{tabular} &
  EN &
  ELI5 &
  \begin{tabular}[c]{@{}c@{}}Web searh\\ results\end{tabular} &
  45k &
  304 &
  104 &
  1.20 &
  auto &
  human \\ \midrule
\begin{tabular}[c]{@{}c@{}}ALCE\\ \citep{Gao2023EnablingLL}\end{tabular} &
  EN &
  \begin{tabular}[c]{@{}c@{}}ASQA\\ QAMPARI, ELI5\end{tabular} &
  \begin{tabular}[c]{@{}c@{}}Wikipedia\\ Common Crawl\end{tabular} &
  3k &
  $$100*k$$ &
  78 &
  N/A &
  N/A &
  auto \\ \bottomrule
\end{tabular}
\caption{Comparison of WebCiteS and relevant datasets. Docs length refers to the total length of the input documents per query. WebCiteS offers two document settings: snippets or full content. The length is measured in characters for Chinese and in words for English. ${\dagger}$ denotes a number reported in the respective paper; otherwise, it's our calculation using open-source data. ALCE limits document length to 100 and varies the number of retrieved documents from 3 to 20. Moreover, it does not offer golden documents for each query and citation annotations. As for the evaluation of citation quality, WebCPM does not consider citation in its evaluation, WebGLM employs human ratings of citation accuracy, and ALCE offers automatic metrics with limitations we seek to address in this work. }
\label{tab:dataset_comp}
\end{table*}

\vspace{-0.5em}
\paragraph{Firstly, most existing datasets are deficient in high-quality citation annotations.}
For instance, the ALCE benchmark~\citep{Gao2023EnablingLL} compiles three question-answering datasets~\citep{fan-etal-2019-eli5,stelmakh-etal-2022-asqa,amouyal2023qampari} without providing citations in the reference answers, limiting its utility for model training. In contrast, WebGLM ~\citep{liu2023webglm} prompts InstructGPT~\citep{ouyang2022training} to generate training data with citations. It controls the citation quality via a sample filtering method which calculates the ROUGE~\cite{lin-2004-rouge} score between the answers and their citations. However, this method focuses on lexical similarity rather than logical entailment, and thus could not precisely measure attribution. 


\paragraph{Secondly, current evaluation methods are insufficient to thoroughly assess attribution.}
Prior works only inspect if the generations are supported by their citations~\cite{Liu2023EvaluatingVI,Gao2023EnablingLL,liu2023webglm} without checking all the documents provided in the input context. However, instances of unsupported generations may result from both \textbf{failing to correctly cite supporting documents} and \textbf{failing to be grounded in all input documents}. Differentiating these two types of errors is crucial for system optimization.
Moreover, existing automatic evaluation~\citep{Gao2023EnablingLL} solely relies on off-the-shelf natural language inference~(NLI) models which only recognize entailment~(full support) and overlook sentences with multiple sub-claims drawing \textbf{partial support} from different sources. 
Such complexities are common in real-world scenarios~\citep{Chen2023ComplexCV,kamoi-etal-2023-wice} and are indicative of a strong capability of synthesizing information across various sources. 

\vspace{0.5em}
To address the above limitations, we present \textbf{WebCiteS}, a Chinese dataset for \textbf{Attributed Query-Focused Summarization~(AQFS)}. As shown in Figure~\ref{fig:intro}, given a query and the retrieved documents, AQFS aims to summarize all pertinent information from the documents with in-line citations to make the summary attributable. 
Our dataset is built upon real-world user queries and search results from \textit{Sogou}, a widely used Chinese web search engine.\footnote{\url{www.sogou.com}} 
We employ human efforts to ensure the quality of summaries and citations. 
Table~\ref{tab:dataset_comp} compares WebCiteS and relevant datasets. 



We propose a comprehensive evaluation framework with a cost-effective automatic evaluator. Our evaluation metrics distinguish two key aspects: \textbf{groundedness}~(if the model outputs are contextually supported) and \textbf{citation quality}~(citation accuracy and comprehensiveness), enabling a more nuanced understanding of attribution errors.
We also train a tailored claim-split model to extract the sub-claims of a sentence for fine-grained verification. This allows the detection of partial support and improves the alignment between our automatic evaluator and human citations. 


Our evaluation of both open-source and proprietary models on WebCiteS reveals the following key findings: 
(1) contextual grounding of generations does not guarantee the avoidance of citation errors, indicating the challenge of explicit attribution in all the tested LLMs; 
(2) although supervised fine-tuning improves both groundedness and citation quality, the top-performing model only reaches a citation F$_1$ score of 76.1\% with about 20\% of sentences not being fully supported by their citations, underscoring the need for further optimization;
(3) models perform worse when summarizing full content of web pages rather than shorter snippets, showing that LLMs are less effective at synthesizing and attributing information in the longer context;
(4) making documents more fine-grained leads to poorer attribution results, highlighting the difficulty LLMs face in pinpointing the exact supporting evidence within the context.

\begin{figure*}[t]
    \centering
    \small
    \includegraphics[width=0.9\linewidth]{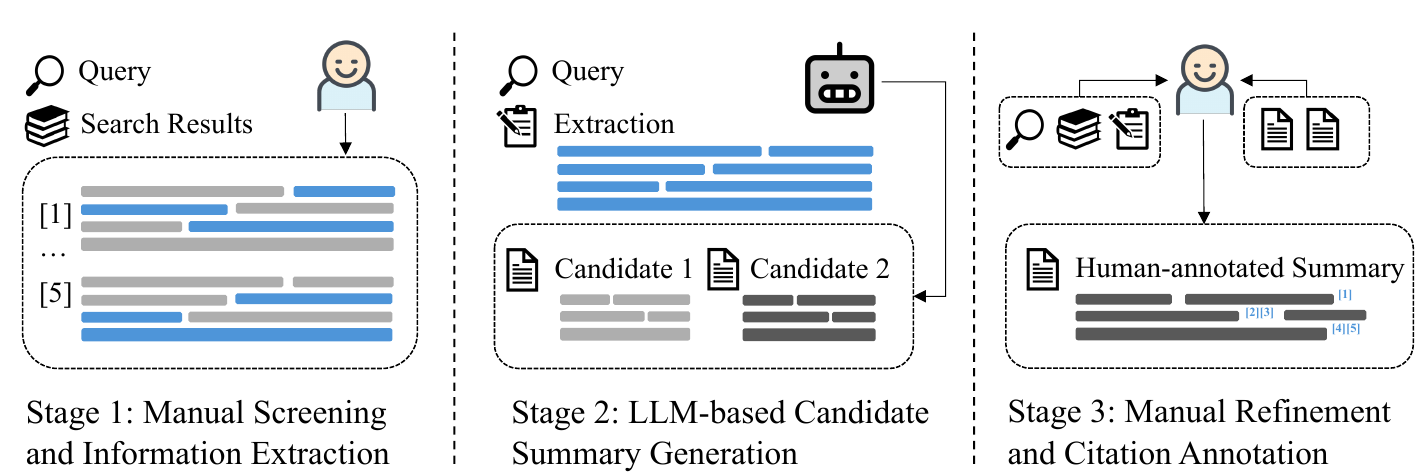}
    \caption{Illustration of the human-LLM collaborative annotation pipeline of WebCiteS. Initially, annotators manually extract useful information from the documents; then, LLMs are used to generate candidate summaries from the extraction; finally, annotators choose the preferred candidate, refine its quality, and annotate citations.}
    \label{fig:annotation}
\end{figure*}

\section{The WebCiteS Dataset}
\label{sec:dataset}
In this section, we first formulate the AQFS task and then delineate the construction of WebCiteS.

\subsection{Task Formulation of AQFS}
\label{subsec:task_formulation}
For a query $q$ and its set of retrieved documents $\mathcal{D}$, AQFS aims to generate a summary $\mathcal{S}$. Following previous works~\cite{Liu2023EvaluatingVI, Gao2023EnablingLL}, we segment it into sentences: $\mathcal{S}=\{s_1,\ldots,s_n\}$. Each sentence $s_i$ cites a subset of documents $\mathcal{C}_i=\{d_{1}, d_{2}, \ldots\}$ where $d_{i}\in \mathcal{D}$. 
Citations are only required for sentences deemed \textit{verification-worthy}~\cite{Liu2023EvaluatingVI}, i.e., sentences needing external evidence for validation. We formulate this property with \textit{citation mask} $\mathcal{M}=\{m_1,...,m_n\}$, where $m_i$ is a binary value and $m_i=1$ denotes sentence $s_i$ requires citations. In practice, we find most of the sentences require citations, except the introductory or concluding sentences in the summary, such as \textit{``There are several possible reasons for papaya tasting a bit bitter''}. 


\subsection{Data Collection}
We collected 40,000 unique, anonymized user queries from \textit{Sogou}, a widely used Chinese web search engine, encapsulating a diverse range of real-world questions. 
After initial refinement, we retained 18,500 non-trivial queries and retrieved five web pages for each query.\footnote{ Common trivial queries include ones that look for word synonyms, text translation, celebrity birth dates, etc. During annotation, we found the top five search results were adequate to address most queries.}
The snippets of web pages returned by the search engine were expanded to a maximum of 250 characters to serve as the documents for the annotation process. 

\subsection{Human-LLM Collaborative Annotation}
\label{subsec:annotation}
Crafting a comprehensive long-form summary from various documents is challenging and labor-intensive for human annotators. Meanwhile, LLMs have showcased impressive proficiency in certain annotation tasks~\cite{He2023AnnoLLMML}. With this in mind, we conceptualized a collaborative annotation pipeline of three stages, as illustrated in Figure~\ref{fig:annotation}.

\paragraph{Stage~1: Manual Screening and Information Extraction.}
Firstly, human annotators read the query and documents thoroughly. They would extract all useful information from the documents and evaluate its utility. We found over 95\% of the queries could be answered by the extracted information, and a few exceptions were discarded.

\paragraph{Stage~2: LLM-based Candidate Summary Generation.}
We leveraged LLMs to construct candidate summaries from the extracted information.
Generating summaries from human-extracted content, as opposed to raw documents, avoided the introduction of irrelevant information. We employed ChatGPT\footnote{We use the gpt-3.5-turbo-0613 checkpoint in this stage.} in the preliminary annotation phase. As our dataset grew, we fine-tuned an open-source model, ChatGLM2-6B~\cite{du2022glm}, to provide an extra candidate summary for each sample.

\paragraph{Stage~3: Manual Refinement and Citation Annotation.}
Lastly, human annotators chose the preferred summary among the two LLM-generated candidates, refined its quality, and annotated citations. 
The chosen summary underwent thorough inspection with non-essential and redundant parts removed. Annotators would cite all supporting documents for verification-worthy sentences, correct groundedness and coherence errors, and supplement missing information. 
Offering multiple candidate summaries aimed to avoid limiting annotators to a single, potentially lower-quality option and to merge the strengths of different options, thereby improving the quality of the final summary. 

\begin{table}[]
\centering
\small
\renewcommand{\arraystretch}{1.1}
\begin{tabular}{@{}l|l@{}}
\toprule
\textbf{Statistic}                           & \textbf{Value}               \\ \midrule
\# Train / Dev / test               & 5,630 / 500 / 1,000 \\
\# Domains                          & 16                  \\
\# Search Results per Query                   & 5                   \\
Full Content Len. / Snippet Len.          & 774 / 205           \\
Summary Len.                     & 167                 \\
\# Sent. per Summary                & 4.56                \\
\# Citations / Sub-claims$^\dagger$ per Sent. & 1.55 / 1.62         \\ \bottomrule
\end{tabular}
\caption{Core statistics of the WebCiteS dataset. Full content length and snippet length refer to the average length of a single search result (web page). $^\dagger$ Sub-claims of sentences are extracted by ChatGPT (Section~\ref{subsec:claimsplit}).}
\label{tab:data_stats}
\end{table}

\paragraph{Quality control.}
We collaborated with crowd-sourcing companies for data labeling. We recruited a team of 27 annotators, 7 quality inspectors, and 1 senior quality inspector, all underwent a month-long training. The quality inspectors reviewed all annotations from the first and third stages, and the senior inspector randomly checked the passed ones. Annotations that failed to meet the standards were returned for corrections and re-inspected.

The core statistics of WebCiteS are present in Table~\ref{tab:data_stats}. Moreover, we conduct the following analysis on the quality of the dataset:

\paragraph{Are the retrieved documents useful to the queries?}
Though we did not ask annotators to explicitly label the relevance score for each document, the human-extracted information from stage 1 could reflect how useful the document is to the query. After annotation, we find that 87.2\% of the documents have human extraction, while the average length of extracted segments per document is 93.4 characters. This indicates that the majority of retrieved documents are helpful to the query.

\paragraph{How much manual refinement is made on candidate summaries?}
The average Levenshtein distance between the human-refined summary and the human-preferred candidate summary is 74.1 (we only count summary edits, excluding citation annotation). This suggests that the quality of candidate summaries were generally judged imperfect by the human annotators.

\paragraph{Overlap of web pages.}

We also notice that previous works~\cite{krishna-etal-2021-hurdles,lewis-etal-2021-question,bolotova-baranova-etal-2023-wikihowqa} point out a high test-train overlap within the dataset would hinder the models to ground generations in the context. While the queries in WebCiteS are non-repetitive, we additionally examine the URLs of all searched web pages serving as documents in different data splits, and find only 0.3\% of the URLs in the train splits exist in validation and test splits, eliminating the concern of high test-train overlap.

See Appendix~\ref{sec:apdx_dataset} for more details of the dataset.


\section{Evaluation Framework}
\label{sec:evaluation}
The AQFS task targets two dimensions: \textbf{summarization utility} and \textbf{attribution}. In this section, we introduce the evaluation metrics based on an evaluator with two key components: a \textbf{claim-split model} $\psi$ and an \textbf{NLI model} $\phi$. The claim-split model decomposes each sentence $s_i$ into a set of sub-claims $\psi(s_i)=\{c_{i,1}, c_{i,2}, \ldots\}$. The NLI model $\phi$ predicts if the given premise entails, contradicts, or is neutral to the given hypothesis. We report the performance of $\phi$ and $\psi$ in Section~\ref{sec:auto_eval}.

\subsection{Evaluating Summarization Utility}
\label{subsec:summarization_metrics}
We adopt the following metrics to evaluate the utility of the summaries:

\paragraph{Length.} We report the average summary length, as prior studies~\citep{10.1613/jair.1.13715,liu-etal-2023-revisiting} point out that the summary lengths across different systems exhibit a large variance which is not well-reflected by other metrics.

\paragraph{Self-BLEU.} Self-BLEU~\citep{zhu18Texygen} measures the diversity of the generated text. \citet{xu-etal-2023-critical} find this metric effective at evaluating the coherence of long-form answers.

\paragraph{Claim precision.} 
We apply $\psi$ to extract all sub-claims from the system summary and calculate the fraction of them being entailed by the reference summary using $\phi$. This metric could measure the \textit{accuracy} and \textit{relevance} of system summaries.

\paragraph{Claim recall.} 
Similarly, we apply $\psi$ to extract all sub-claims from the reference summary and calculate the fraction of them being entailed by the system summary. This metric could measure the \textit{comprehensiveness} of system summaries.

\paragraph{Claim $\text{F}_1$.}
Finally, we can compute the claim $\text{F}_1$ score of a system by taking the harmonic mean of its claim precision and recall. See Appendix~\ref{subsec:apdx_summ_metrics} for more discussions on summarization metrics.

\subsection{Evaluating Attribution}
\label{subsec:attribution_metrics}
Only \textit{verification-worthy} sentences with \textit{citation mask} $m_i=1$ are included in attribution evaluation. \citeposs{Gao2023EnablingLL} automatic metrics assume that all sentences need citations, i.e. the citation mask $m_i$ is always 1.
However, we observe some exceptions such as the introductory or concluding sentences in the summaries~(see Section~\ref{subsec:task_formulation}). 
Therefore, we propose to automatically predict the citation mask for sentence $s_i$:
$$
m_i=\mathds{1}(\mathcal{C}_i \neq \emptyset \vee  \phi(\mathcal{S}^*,s_i)\neq \text{entailment})
$$
where $s_i \in \mathcal{S}$, $\mathcal{S}^* =\{s_j|s_j\in \mathcal{S}-\{s_i\}, \mathcal{C}_j \neq \emptyset \}$
Namely, the citation mask $m_i$ is 1 if one of the conditions is satisfied: (1) The citations $\mathcal{C}_i$ of $s_i$ is not empty, as all model-generated citations should be verified; (2) $s_i$ is not entailed by $\mathcal{S}^*$, as we assume introductory or concluding sentences should be entailed by the rest of the summary.\footnote{$\mathcal{S}^*$ only involves sentences with non-empty citations. Without this restriction, the evaluation may be hacked if the model generates two mutually entailed sentences without citations. In this case, both of their citation masks would be 0, making them escape from attribution evaluation.}

For sentences with citation mask $m_i=1$, we evaluate two aspects for attribution: \textbf{groundedness} and \textbf{citation quality} with the following metrics:

\paragraph{AIS.} 
The AIS score assesses if the generation is \textit{attributable to identified sources}~\citep{Bohnet2022AttributedQA,Rashkin2023attr}. We adopt the fine-grained version proposed in RARR~\citep{gao-etal-2023-rarr}, which calculates the fraction of attributable sentences in the generation. 
Since citations serve as the \textit{identified sources} in AQFS, \textbf{a sentence $s_i$ is attributable if it is fully supported by its citations} $\mathcal{C}_i$. In practice, our automatic evaluator will classify $s_i$ as attributable if (1) $s_i$ does not contradict any citation in $\mathcal{C}_i$ and (2) $s_i$ or \textbf{all} of its sub-claims $\psi(s_i)$ are entailed by the citations. 
Under this definition, the AIS score is equivalent to the \textit{citation recall} metric proposed in ALCE~\citep{Gao2023EnablingLL}. Since this metric generally measures both citation quality and groundedness, we adopt the term \textit{AIS} and define a variant of \textit{citation recall}. 

\paragraph{ACS.} 
We propose \textit{attributable to contextual sources} (ACS), a variant of the AIS score that uses oracle citations from the evaluator rather than the model-generated citations for evaluation. This isolates groundedness assessment by eliminating the impact of citation errors. For example, if $s_i$ is contextually grounded but does not cite any source, its AIS and ACS scores will be 0 and 1 respectively.

\paragraph{Citation precision.}
This metric measures if the sentence is correctly supported by each of its citations. For $s_i$, we extract the model-generated citations $C^i_\text{pred}$ and obtain the oracle citations $C^i_\text{ref}$ involving all $d_j \in \mathcal{D}$ that \textbf{fully or partially support} $s_i$. 
Then, citation precision for $s_i$ is:
$$
\text{Citation Prec}(s_i) =
\frac{\lvert C_{\text{pred}}^i \cap C_{\text{ref}}^i \rvert}{\lvert C_{\text{pred}}^i \rvert}
$$
In practice, the evaluator would include $d_j$ into $C^i_\text{ref}$ if 
(1) it entails $s_i$ or
(2) it does not contradict $s_i$ and entails \textbf{any} of its sub-claims.
Moreover, if $C^i_\text{pred}$ is empty, we seek the nearest non-empty citations $C^{i*}_{\text{Pred}}$ from the subsequent sentences of $s_i$ to replace $C_{\text{Pred}}^i$ for evaluation. This is to avoid penalizing the model for generating multiple sentences based on the same sources but only adding citations in the final sentence. If $C^{i*}_{\text{Pred}}$ is not found, citation precision for $s_i$ woule be zero. Finally, we average the scores of all $s_i \in \mathcal{S}, m_i=1$ as the citation precision score for the summary $\mathcal{S}$.

\paragraph{Citation recall.}
This metric measures if the sentence comprehensively cites all supporting sources:
$$
\text{Citation Rec}(s_i) =
\frac{\lvert C_{\text{pred}}^i \cap C_{\text{ref}}^i \rvert}{\lvert C_{\text{ref}}^i \rvert}
$$
Similarly, we seek $C^{i*}_{\text{Pred}}$ if $C^i_\text{pred}$ is empty. We assign a zero score if $C^{i}_{\text{Ref}}$ is empty. Finally, we average the citation recall scores of all $s_i \in \mathcal{S}, m_i=1$.

\paragraph{Citation $\text{F}_1$.}
Similar to Claim $\text{F}_1$, we compute the citation $\text{F}_1$ score of a system by taking the harmonic mean of its citation precision and recall. 

\begin{figure}[t]
    \centering
    \vspace{-1em}
    \includegraphics[width=\linewidth]{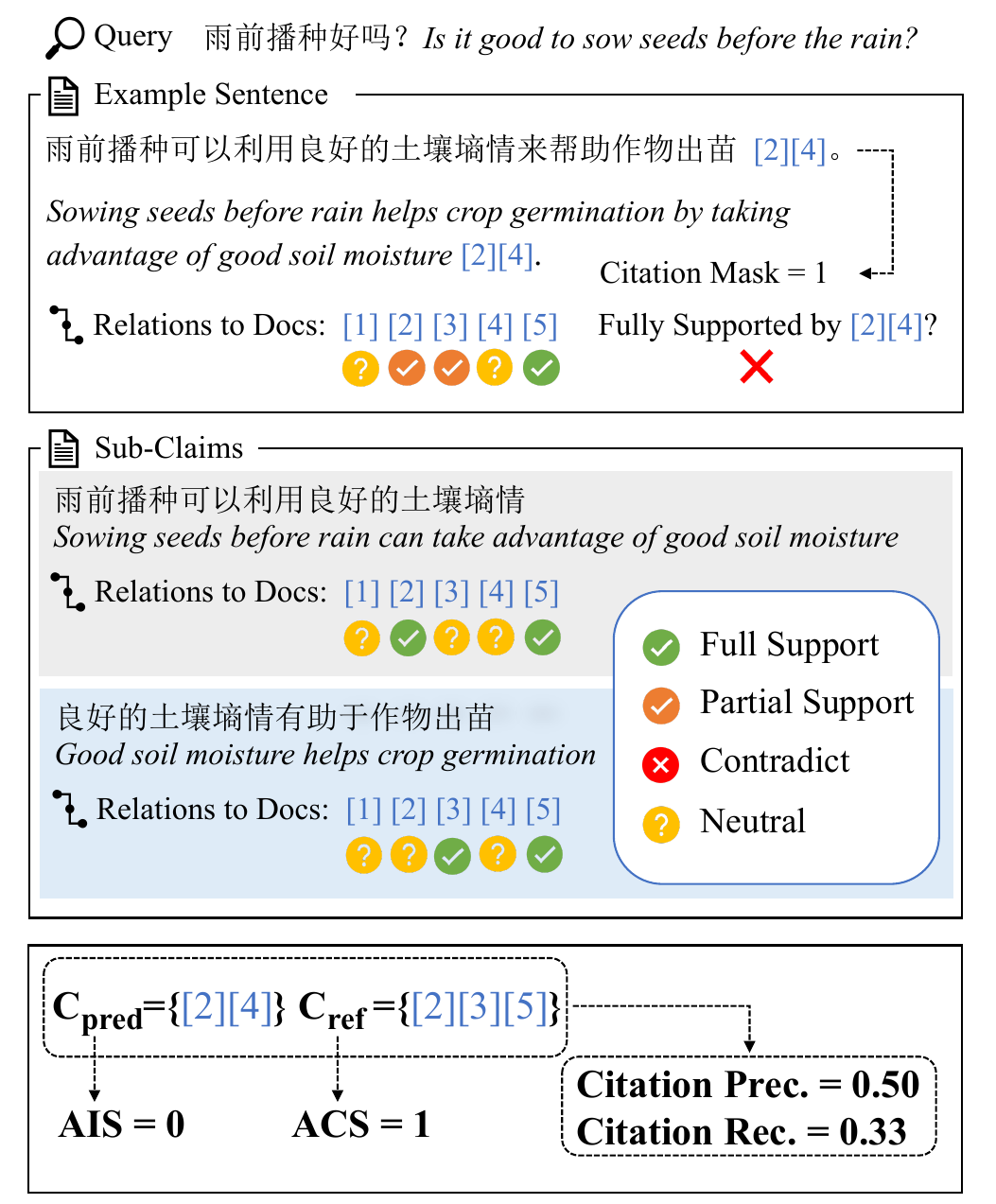}
    \caption{Illustration of our attribution evaluation. We use a claim-split model to extract sub-claims of a sentence and conduct fine-grained verification on all the source documents. The translation is in italic text.}
    \label{fig:evaluation}
\end{figure}

Figure~\ref{fig:evaluation} shows the framework of attribution evaluation. Compared to \citeposs{Gao2023EnablingLL} automatic methods, our method (1) considers citation mask, (2) incorporates the claim-split model for partial support detection, and (3) distinguishes groundedness and citation quality.
See Appendix~\ref{subsec:apdx_attr_metrics} for more discussions on attribution metrics.

\section{Evaluating the Automatic Evaluator}
\label{sec:auto_eval}
In this section, we assess the reliability of the automatic evaluator with $\phi$ and $\psi$.

\subsection{Performance of the NLI model}
We evaluate the performance of different open-source NLI models in predicting human-annotated citations in WebCiteS. We finally choose an mT5 model~\citep{xue-etal-2021-mt5} fine-tuned on multilingual NLI tasks as $\phi$ for our evaluator,\footnote{\url{huggingface.co/alan-turing-institute/mt5-large-finetuned-mnli-xtreme-xnli}} since it achieves the highest accuracy of 82.3\% among all models. See Appendix~\ref{sec:apdx_nli} for more details about NLI models.

\subsection{Performance of the Claim-Split Model}
\label{subsec:claimsplit}
We first prompt ChatGPT\footnote{We use the gpt-3.5-turbo-1106 checkpoint.} to extract the sub-claims in sentences since \citet{kamoi-etal-2023-wice} have validated this approach via human assessment.
As using proprietary LLMs for automatic evaluation still faces limitations in efficiency and cost, we additionally fine-tune mT5 on the outputs of ChatGPT to learn this task.
We evaluate the claim-split models with the following metrics:

\paragraph{Redundancy.} It measures if the model generates redundant sub-claims of the source sentence. Two sub-claims $c_i$ and $c_j$ in $\psi(s)$ are deemed redundant if they entail each other: $\phi(c_i,c_j)=1$ and $\phi(c_j,c_i)=1$. 
Based on this, we could eliminate the redundancy of $\psi(s)$: if multiple sub-claims are redundant, we only keep the first one and remove the rest. The resulting set is denoted as $\psi^*(s)$. Finally, the metric is computed as: 
$$
\text{Redundancy}(\psi(s)) = \frac{\rvert\psi(s)\rvert - \rvert\psi^*(s)\rvert}{\rvert\psi(s)\rvert}
$$

\paragraph{\#~Splits.} It is defined as the average count of non-redundant sub-claims per sentence, which could reflect the granularity of model outputs, as a lower value indicates that the model may not effectively separate some sub-claims in the source sentence.

\paragraph{Correctness.} It is defined as the fraction of sub-claims being entailed by the source sentence using the NLI model $\phi$, as we assume that all correct sub-claims should be entailed by the source sentence.

\paragraph{Completeness.} It is a binary value measuring if the source sentence is entailed by the concatenation of all sub-claims using the NLI model $\phi$. If not, some essential sub-claims may be missing in the model outputs.

\vspace{0.5em}
Table~\ref{tab:claimsplit} shows the evaluation result. We first notice that ChatGPT and the fine-tuned models are consistent in \textit{\# splits} which reflects the decomposition granularity. Moreover, the fine-tuned models slightly outperform ChatGPT in \textit{correctness} and \textit{completeness}.
The primary reason is that ChatGPT occasionally unfollow the task instruction~\citep{zhou2023instructionfollowing}.
Therefore, we select the fine-tuned mT5-large model as $\psi$ for our evaluator. See Appendix~\ref{sec:apdx_claimsplit} for more details about claim-split models.

\begin{table}
\centering
\small
\setlength\tabcolsep{3pt}
\renewcommand{\arraystretch}{1.1}
\begin{tabular}{@{}ccccc@{}}
\toprule
   & \textbf{Redun.} ↓   & \textbf{\# Splits}     & \textbf{Correct.}    &  \textbf{Complete.}   \\ \midrule
mT5-Base     & 2.4          & \textbf{1.9}          & \textbf{99.0} & 90.0          \\
mT5-Large    & \textbf{1.7}  & \textbf{1.9}          & \textbf{99.0}          & \textbf{92.2} \\
ChatGPT & \textbf{1.7} & \textbf{1.9} & 96.7          & 89.3          \\ \bottomrule
\end{tabular}
\caption{Performance of different claim-split models. }
\label{tab:claimsplit}
\end{table}

\begin{table}
\centering
\small
\setlength\tabcolsep{2pt}
\renewcommand{\arraystretch}{1.1}
\begin{tabular}{@{}ccccc@{}}
\toprule
\textbf{Claim-Split} &
  \begin{tabular}[c]{@{}c@{}}\textbf{Citation}\\ \textbf{Prec.}\end{tabular} &
  \begin{tabular}[c]{@{}c@{}}\textbf{Citation}\\ \textbf{Rec.}\end{tabular} &
  \begin{tabular}[c]{@{}c@{}}\textbf{AIS}\\ \textbf{(HumanCite)}\end{tabular} &
  \begin{tabular}[c]{@{}c@{}}\textbf{AIS}\\ \textbf{(AutoCite)}\end{tabular} \\ \midrule
\multicolumn{5}{c}{\cellcolor[HTML]{EFEFEF}\textbf{Default Citation Mask}}             \\
$-$         & 77.7          & 74.5          & 84.8          & 90.3          \\
mT5-Large & 77.7          & 82.3           & 87.3           & 93.1          \\
ChatGPT   & 77.9          & 81.0          & 87.1          & 92.7          \\
\multicolumn{5}{c}{\cellcolor[HTML]{EFEFEF}\textbf{Auto Citation Mask}}             \\
$-$         & 81.7           & 73.4          & 84.8          & 89.8          \\
mT5-Large & 81.7          & 82.3          & 87.4          & 92.7          \\
ChatGPT   & 81.9          & 80.7          & 87.1          & 92.4          \\
\multicolumn{5}{c}{\cellcolor[HTML]{EFEFEF}\textbf{Human Citation Mask}}              \\
$-$         & 82.4          & 74.0        & 85.4          & 90.0          \\
mT5-Large & 82.4          & \textbf{83.0} & \textbf{88.2} & \textbf{93.1} \\
ChatGPT   & \textbf{82.6} & 81.3          & 87.9          & 92.7          \\ \bottomrule
\end{tabular}
\caption{Performance of the automatic evaluator on evaluating the attribution in human-annotated summaries. We use a fixed NLI model and vary the use of claim-split models under different citation mask settings.}
\label{tab:auto_citation_eval}
\end{table}

\subsection{Performance of the Automatic Evaluator}
\label{subsec:test_auto_eval}
 Finally, we assess our automatic evaluator with $\phi$ and $\psi$ on the test set of WebCiteS. We use it to predict the citations for sentences in the reference summaries, and then assess those citations by taking human citations as ground truth. We also compute the AIS scores using both citations. 
 We use the same NLI model $\phi$ and vary the use of $\psi$ to analyze the impact of the claim-split strategy.
 Besides, we compare different citation mask settings: 
 (1) \textit{default}, which sets $m_i=1$ for all sentences; 
 (2) \textit{auto}, which automatically predicts $m_i$ (see Section~\ref{subsec:attribution_metrics});
 (3) \textit{human}, which only sets $m_i=1$ for sentences with human citations. The results in Table~\ref{tab:auto_citation_eval} show that: 
 \textbf{(1) Claim-split model helps to detect partial support.} Integrating $\psi$ enhances both citation recall and AIS scores, indicating that the citations that only support part of the sub-claims of the sentences are effectively identified; 
 \textbf{(2) Accurate citation mask improves the performance of the evaluator.} Using the human citation mask yields the best overall performance, while the auto citation mask achieves better citation precision compared to the default setting. This result emphasizes the necessity of identifying if a sentence requires citations during evaluation. 
 
Moreover, the Cohen’s kappa coefficient between the evaluator (using mT5-large as $\psi$ and auto citation mask) and human annotators on whether a sentence should cite a document is 0.6483, which suggests substantial agreement. This further validates the reliability of the automatic evaluator.

\begin{table*}[t]
\centering
\small
\vspace{-0.5em}
\renewcommand{\arraystretch}{1.2}
\begin{tabular}{@{}cccccccccccc@{}}
\toprule
                     &                    & \multirow{2}{*}{\textbf{Len.}} & \textbf{Self-}        & \multicolumn{3}{c}{\textbf{Claim}}                     & \multicolumn{3}{c}{\textbf{Citation}}                  & \multirow{2}{*}{\textbf{AIS}} & \multirow{2}{*}{\textbf{ACS}} \\
                     &                    &                       & \textbf{BLEU} ↓       & \textbf{Prec.}         & \textbf{Rec.}          & \textbf{F$_1$}            & \textbf{Prec.}         & \textbf{Rec.}          & \textbf{F$_1$}            &                      &                      \\ \midrule
\multirow{6}{*}{FSP} & ChatGPT            & 223                   & 12.2         & 53.5          & 56.5          & \textbf{54.9} & 71.2          & 64.8          & 67.8          & 75.1                 & 84.7             \\
                     & GPT-4              & 285                   & 5.6          & 46.1          & \textbf{65.8} & 54.2          & 72.2          & 71.4          & 71.8          & 75.7                 & 81.1             \\ 
                     & ChatGLM2-6B        & 256                   & 11.7         & 40.1          & 45.1          & 42.5          & 9.9           & 7.3           & 8.4           & 9.8                  & 73.4            \\
                     & ChatGLM3-6B        & 252                   & 10.0         & 45.6          & 53.7          & 49.3          & 57.0          & 51.4          & 54.1          & 60.0                 & 85.0            \\
                     & Baichuan2-7B  & 188                   & 7.6          & 46.9          & 48.4          & 47.6          & 16.3          & 21.6          & 18.6          & 20.8                 & 81.0            \\
                     & Baichuan2-13B & 200                   & 7.8          & 45.4          & 49.6          & 47.4          & 44.4          & 51.0          & 47.4          & 52.3                 & 76.2           \\ \midrule
\multirow{6}{*}{SFT} & mT5-Base           & 142                   & 19.7         & 47.2          & 35.9          & 40.8          & 47.2          & 35.9          & 37.7          & 40.0                 & 82.0            \\
                     & mT5-Large          & 142                   & 15.7         & 52.9          & 40.6          & 45.9          & 50.2          & 51.1          & 50.6          & 56.0                 & \textbf{90.3}   \\
                     & ChatGLM2-6B        & 156                   & 8.7          & 54.6          & 47.6          & 50.9          & 76.3          & 72.5          & 74.4          & 79.4                 & 87.1             \\
                     & ChatGLM3-6B        & 163                   & 8.9          & 55.7          & 49.3          & 52.3          & \textbf{78.5} & \textbf{73.8} & \textbf{76.1} & \textbf{81.3}        & 89.4             \\
                     & Baichuan2-7B  & 212                   & 9.9          & 52.4          & 53.1          & 52.8          & 68.1          & 67.8          & 67.9          & 71.6                 & 81.7            \\
                     & Baichuan2-13B & 132                   & \textbf{5.3} & \textbf{58.9} & 42.1          & 49.1          & 67.7          & 72.0          & 69.8          & 74.4                 & 81.0       \\     
\bottomrule
\end{tabular}
\caption{Results of the AQFS task on WebCiteS, where each sample consists of five snippets as documents.}
\label{tab:main_results}
\end{table*}

\section{Experiments on the AQFS Task}
\label{sec:main_exp}
We evaluate various models on the AQFS task via two methods: \textbf{few-shot prompting (FSP)} and \textbf{supervised fine-tuning (SFT).} Our prompt for FSP consists of the task instruction and one-shot demonstration, while the SFT prompt removes the demonstration and condenses the instruction for efficiency.
For open-source models, we evaluate mT5, ChatGLM2-6B, ChatGLM3-6B~\citep{du2022glm}, Baichuan2-7B, and Baichuan2-13B~\citep{Yang2023Baichuan2O} via both FSP and SFT.\footnote{We use the \textit{chat} version of Baichuan2 models.}
For proprietary models, we evaluate ChatGPT and GPT-4 via FSP.\footnote{We use the gpt-3.5-turbo-1106 checkpoint for ChatGPT and the gpt-4-1106-preview checkpoint for GPT-4.} See Appendix~\ref{subsec:apdx_aqfs_implement} for implementation details.

\subsection{Main Results}
\label{sebsec:main_result}
We first adopt the default setting where each sample consists of five snippets of web pages as documents.
The experimental results are present in Table~\ref{tab:main_results}. In general, we observe a large variance of summary lengths across models. For FSP, ChatGPT and GPT-4 outperform all open-source LLMs on both claim F$_1$ scores and citation F$_1$ scores. However, even GPT-4 exhibits unignorable attribution errors: only 72\% of its citations are correct and only 71\% of supporting documents are cited. Moreover, only 76\% of its generated sentences are fully supported by their respective citations, and only 81\% of them are grounded in the input context. For SFT, we observe that smaller pre-trained models such as mT5 significantly lag behind open-source LLMs on both summarization utility and citation quality. Although the fine-tuned mT5-Large model achieves the best groundedness (reflected by the highest ACS score of 90.3\%), we find it is primarily due to the model's tendency to copy the input text rather than summarize the content pertinent to the query, which leads to the suboptimal claim F$_1$ scores. Our additional findings include:


\vspace{-0.6em}
\paragraph{Groundedness errors and citation errors coexist across models.} No model achieves a perfect ACS score, indicating the presence of groundedness errors. Moreover, all models' ACS scores exceed the respective AIS scores. Since their gaps are simply brought by citation errors, this finding reveals that even if the model grounds its generation contextually, it could still struggle with accurate citations. This underscores the challenge of explicit attribution in both open-source and proprietary LLMs.

\vspace{-0.6em}
\paragraph{Supervised fine-tuning improves attribution.} Without fine-tuning, all open-source LLMs struggle with accurate citations. However, SFT consistently boosts all attribution metrics and narrows the gaps between the AIS and ACS scores, indicating a simultaneous optimization of both groundedness and citation quality. This finding highlights the potential benefits of involving the AQFS task during instruction-tuning~\citep{zhang2023instruction} to enhance attribution in open-source LLMs.

\begin{table}[]
\centering
\small
\setlength\tabcolsep{1.5pt}
\renewcommand{\arraystretch}{1.1}
\vspace{-1.5em}
\begin{tabular}{@{}ccccccc@{}}
\toprule
\textbf{\begin{tabular}[c]{@{}c@{}}Source\\ Type\end{tabular}} & \textbf{\begin{tabular}[c]{@{}c@{}}Max Doc \\ Length\end{tabular}} & \textbf{\# Docs} & \textbf{\begin{tabular}[c]{@{}c@{}}Claim\\ F$_1$\end{tabular}} & \textbf{\begin{tabular}[c]{@{}c@{}}Citation\\ F$_1$\end{tabular}} & \textbf{AIS}  & \textbf{ACS}  \\ \midrule
\multicolumn{7}{c}{\cellcolor[HTML]{EFEFEF}\textbf{ChatGPT}}                                                                                                                                                                                                                                                         \\
Snippet                                                       & 250                                                                & 5                & \textbf{54.9}                                               & 67.8                                                           & 75.1          & 84.7          \\
Full Content                                                  & 512                                                                & 11               & 45.9                                                        & 58.2                                                           & 72.2          & 88.9          \\
Full Content                                                  & 256                                                                & 20               & 44.9                                                        & 51.3                                                           & 65.2          & 86.5          \\
\multicolumn{7}{c}{\cellcolor[HTML]{EFEFEF}\textbf{ChatGLM3-6B (SFT)}}                                                                                                                                                                                                                                                     \\
Snippet                                                       & 250                                                                & 5                & 52.3                                                        & \textbf{76.1}                                                  & \textbf{81.3} & \textbf{89.4} \\
Full Content                                                  & 512                                                                & 11               & 43.6                                                        & 51.9                                                           & 65.8          & 88.3          \\
Full Content                                                  & 256                                                                & 20               & 41.8                                                        & 42.9                                                           & 56.3          & 84.9          \\
\multicolumn{7}{c}{\cellcolor[HTML]{EFEFEF}\textbf{Baichuan2-7B (SFT)}}                                                                                                                                                                                                                                                    \\
Snippet                                                       & 250                                                                & 5                & 52.8                                                        & 67.9                                                           & 71.6          & 81.7          \\
Full Content                                                  & 512                                                                & 11               & 41.3                                                        & 51.3                                                           & 63.7          & 83.2          \\
Full Content                                                  & 256                                                                & 20               & 41.1                                                        & 42.8                                                           & 53.9          & 80.5          \\ \bottomrule
\end{tabular}
\caption{Impact of different document settings. Besides the default setting in Table~\ref{tab:main_results}, we further evaluate the model performance in summarizing the full content of web pages. We also adopt different chunk sizes (512 and 256) to analyze the impact of document granularity.}
\label{tab:fulldoc}
\end{table}

\subsection{Results in the Long-Context Setting}
\label{subsec:long_context}
We further adopt a more challenging long-context setting, where the models are provided with the full content of web pages to summarize. We chunk the web pages into passages with a maximum length of 512 or 256 and assign a unique citation number to each of them.\footnote{Our evaluator is based on the mT5 model which supports a context window of 512 tokens.} Table~\ref{tab:fulldoc} presents the performance of the selected models, which shows that:

\begin{figure}
    \centering
    \vspace{-1.5em}
    \includegraphics[width=\linewidth]{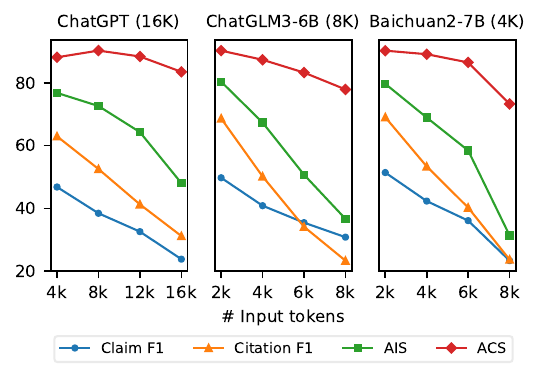}
    \caption{Performance change over context length of the models in Table~\ref{tab:fulldoc}, where full content of web pages are chunked into documents with a maximum length of 512. Model names are followed by their context window size. The number of input tokens is counted using the tokenizer of each model respectively.}
    \label{fig:fulldoc}
\end{figure}

\vspace{-0.5em}
\paragraph{Extending context length reduces model performance.}
We observe an overall decline in both summarization utility and attribution with full content instead of snippets. We also visualize the performance variance over the context length in Figure~\ref{fig:fulldoc}, which shows a decline in claim F$_1$, citation F$_1$, and AIS scores as the context length extends. This pattern indicates that extending context length challenges models' ability to synthesize pertinent information and correctly cite sources.

\vspace{-0.5em}
\paragraph{Using fine-grained documents poses challenges in attribution.}
In practice, if the cited documents are too long, it is challenging to verify the model generations, and we expect the models could cite more specific segments of information. Therefore, we also vary the maximum document length when chunking the web pages and investigate the impact of document (citation) granularity. As shown in Table~\ref{tab:fulldoc}, reducing max document length from 512 to 256, without changing the total input evidence, drastically lowers citation F$_1$ and AIS scores for all three models.
This performance reduction reveals the difficulty LLMs face in precisely pinpointing the exact supporting evidence within the context.

\section{Related Work}
\label{sec:related_work}
\paragraph{Relevant datasets and benchmarks.}
Query-focused multi-document summarization (QF-MDS) aims to summarize multiple documents driven by specific queries~\citep{tombros1998advantages,li-li-2013-novel,qfmdsreview}. Similarly, long-form question answering (LFQA) focuses on producing detailed answers, often utilizing external sources~\citep{krishna-etal-2021-hurdles}.  
Most existing datasets in both tasks do not consider attribution in the setup~\citep{fan-etal-2019-eli5,fabbri-etal-2019-multi,boni2021howsumm,stelmakh-etal-2022-asqa,bolotova-baranova-etal-2023-wikihowqa}.
\citet{Bohnet2022AttributedQA} propose the attributed question answering~(AQA) benchmark where the system must output the answer alongside a piece of evidence. However, long-form responses usually require citing multiple sources of evidence.
Recent initiatives~\citep{qin2023webcpm,liu2023webglm,Gao2023EnablingLL} in this dimension exhibit limitations in citation annotation and evaluation methods detailed in Section~\ref{sec:intro} and Table~\ref{tab:dataset_comp}.

\paragraph{Evaluating attribution in LLMs.}
Attribution refers to the ability to provide external evidence supporting the claims made by the model \citep{Rashkin2023attr,li2023survey}. It is crucial for enhancing the credibility of generations and reducing hallucinations~\citep{10.1145/3571730, zhang2023sirens}. Although attribution can be approached through various methods, such as generating references from the model's internal knowledge~\citep{weller2023according}, retrieval-augmented generation~(RAG) with citations~\citep{Nakano2021WebGPTBQ,qin2023webcpm,liu2023webglm,Gao2023EnablingLL,li2023llatrieval}, and seeking references post-generation~\cite{gao-etal-2023-rarr,Chen2023ComplexCV,10.1145/3624918.3625336}, cost-effective evaluation methods remain a challenging task. Existing automatic metrics~\cite{Gao2023EnablingLL,Bohnet2022AttributedQA,Yue2023AutomaticAttr} solely depend on off-the-shelf NLI models, failing to detect partial support over complex claims. On the other hand, recent works on summarization evaluation~\cite{liu-etal-2023-revisiting}, textual entailment~\cite{kamoi-etal-2023-wice}, and fact verification~\cite{Chen2023ComplexCV,min-etal-2023-factscore} leverage claim decomposition strategy for fine-grained verification. However, they heavily rely on either human efforts or proprietary LLMs to extract sub-claims, which restricts their scalability. Moreover, existing works do not distinguish the evaluation of groundedness and citation quality. This limits the in-depth understanding of attribution errors in different models.

\section{Conclusion}
\label{sec:conclusion}
In this work, we formulate the task of attributed query-focused summarization (AQFS) and present WebCiteS, a high-quality dataset derived from real-world user queries and search results. 
We propose a comprehensive evaluation framework on summarization utility and attribution. Notably, we highlight two fine-grained dimensions of attribution: groundedness and citation quality.
We further enhance the framework with a carefully-designed automatic evaluator, and validate its substantial agreement with human annotators.
Finally, we evaluate both open-source and proprietary LLMs extensively on WebCiteS to underscore the unsolved challenge of attribution, especially in the long-context setting with fine-grained documents. We believe WebCiteS could facilitate more future explorations in attributed language models.

\section*{Limitations}
\paragraph{Task Setup and Dataset.}
The AQFS task and the WebCiteS dataset primarily focus on evaluating and improving the model's ability to synthesize information from multiple sources with accurate attribution. Therefore, we do not incorporate retrieval into our task setup. Though we find most web pages returned by the search engine are relevant to the queries (Section~\ref{subsec:annotation}), other works highlight the importance of retrieval quality~\citep{Gao2023EnablingLL,liu2023webglm}. We believe the search results and snippets provided in WebCiteS can also serve as a valuable resource to refine open-source retrievers in future works.

\paragraph{Evaluation.}
The reliability of our automatic evaluator is dependent on the accuracy of both the claim-split model and the NLI model. Though we have validated their performance in Section~\ref{sec:auto_eval}, we could not ensure the model-generated sub-claims are atomic in granularity. Failing to divide sub-claims could affect the identification of partial support instances. Moreover, the context window of the NLI model is also a constraint. Though the mT5 model uses relative position embeddings~\cite{shaw-etal-2018-self} and accepts arbitrary input sequence length, we find its accuracy drops if the input sequence length significantly exceeds its context window of 512 tokens, primarily due to the length distribution of its training data.
Therefore, we believe training a more reliable NLI model for the long-context setting is also an important future work.

\section*{Ethics Statement}
Since WebCiteS is built upon real user queries, we have taken strict measures to address privacy issues. We only sampled anonymized queries from the search log without collecting any other information such as user identifiers. All queries were shuffled and not present in time order, making it impossible to obtain individual search history from the dataset. Lastly, during annotation, we asked annotators and quality inspectors to pay attention to discard any query with potential privacy issues.

Moreover, we have endeavored to eliminate all inappropriate content from our corpus. Firstly, we adopt an internal commercial tool to automatically detect and discard queries with improper intentions. Secondly, the commercial search engine used in this work has taken content quality and safety into account during web page retrieval and ranking, so it is unlikely that the top five search results would contain dubious or harmful material. Thirdly, human annotators were asked to discard any samples with inappropriate content (see Appendix~\ref{sec:apdx_dataset}). Such a manual inspection process, despite being essential to enhance the dataset quality, would inevitably expose potential risks to the annotators themselves. We attempted to minimize those risks by automatically filtering most inappropriate content with the commercial tool and search engine before annotation. In practice, of all samples discarded by annotators, only less than 5\% fell into the category of inappropriate content. Meanwhile, we alerted annotators of the potential risks in advance by including a cautionary note in the annotation instruction and allowed them to skip any sample that made them uncomfortable.

Finally, this work focuses on \textit{attribution} rather than \textit{factuality}: even if a response is fully supported by the evidence it attributes, it is not guaranteed to be factual since the evidence itself might contain factual errors or become outdated. Just as previous works~\citep{Bohnet2022AttributedQA, Rashkin2023attr} point out, the judge of factuality, especially in open domains, is extremely difficult. During the construction of WebCiteS, though we requested annotators to discard samples with highly questionable materials, we did not assume that their professional expertise could cover all fields in the corpus. Therefore, we emphasize that future works should be cautious about treating the annotated summaries in WebCiteS as "facts".

\bibliography{custom}

\begin{thebibliography}{57}
\expandafter\ifx\csname natexlab\endcsname\relax\def\natexlab#1{#1}\fi

\bibitem[{Amouyal et~al.(2023)Amouyal, Wolfson, Rubin, Yoran, Herzig, and Berant}]{amouyal2023qampari}
Samuel~Joseph Amouyal, Tomer Wolfson, Ohad Rubin, Ori Yoran, Jonathan Herzig, and Jonathan Berant. 2023.
\newblock \href {http://arxiv.org/abs/2205.12665} {Qampari: An open-domain question answering benchmark for questions with many answers from multiple paragraphs}.

\bibitem[{Bohnet et~al.(2022)Bohnet, Tran, Verga, Aharoni, Andor, Soares, Eisenstein, Ganchev, Herzig, Hui, Kwiatkowski, Ma, Ni, Schuster, Cohen, Collins, Das, Metzler, Petrov, and Webster}]{Bohnet2022AttributedQA}
Bernd Bohnet, Vinh~Q. Tran, Pat Verga, Roee Aharoni, Daniel Andor, Livio~Baldini Soares, Jacob Eisenstein, Kuzman Ganchev, Jonathan Herzig, Kai Hui, Tom Kwiatkowski, Ji~Ma, Jianmo Ni, Tal Schuster, William~W. Cohen, Michael Collins, Dipanjan Das, Donald Metzler, Slav Petrov, and Kellie Webster. 2022.
\newblock \href {https://arxiv.org/abs/2212.08037} {Attributed question answering: Evaluation and modeling for attributed large language models}.
\newblock \emph{ArXiv preprint}, abs/2212.08037.

\bibitem[{Bolotova et~al.(2023)Bolotova, Blinov, Filippova, Scholer, and Sanderson}]{bolotova-baranova-etal-2023-wikihowqa}
Valeriia Bolotova, Vladislav Blinov, Sofya Filippova, Falk Scholer, and Mark Sanderson. 2023.
\newblock \href {https://doi.org/10.18653/v1/2023.acl-long.290} {{W}iki{H}ow{QA}: A comprehensive benchmark for multi-document non-factoid question answering}.
\newblock In \emph{Proceedings of the 61st Annual Meeting of the Association for Computational Linguistics (Volume 1: Long Papers)}, pages 5291--5314, Toronto, Canada. Association for Computational Linguistics.

\bibitem[{Boni et~al.(2021)Boni, Feigenblat, Lev, Shmueli-Scheuer, Sznajder, and Konopnicki}]{boni2021howsumm}
Odellia Boni, Guy Feigenblat, Guy Lev, Michal Shmueli-Scheuer, Benjamin Sznajder, and David Konopnicki. 2021.
\newblock \href {https://arxiv.org/abs/2110.03179} {Howsumm: a multi-document summarization dataset derived from wikihow articles}.
\newblock \emph{ArXiv preprint}, abs/2110.03179.

\bibitem[{Brown et~al.(2020)Brown, Mann, Ryder, Subbiah, Kaplan, Dhariwal, Neelakantan, Shyam, Sastry, Askell, Agarwal, Herbert{-}Voss, Krueger, Henighan, Child, Ramesh, Ziegler, Wu, Winter, Hesse, Chen, Sigler, Litwin, Gray, Chess, Clark, Berner, McCandlish, Radford, Sutskever, and Amodei}]{brown2020language}
Tom~B. Brown, Benjamin Mann, Nick Ryder, Melanie Subbiah, Jared Kaplan, Prafulla Dhariwal, Arvind Neelakantan, Pranav Shyam, Girish Sastry, Amanda Askell, Sandhini Agarwal, Ariel Herbert{-}Voss, Gretchen Krueger, Tom Henighan, Rewon Child, Aditya Ramesh, Daniel~M. Ziegler, Jeffrey Wu, Clemens Winter, Christopher Hesse, Mark Chen, Eric Sigler, Mateusz Litwin, Scott Gray, Benjamin Chess, Jack Clark, Christopher Berner, Sam McCandlish, Alec Radford, Ilya Sutskever, and Dario Amodei. 2020.
\newblock \href {https://proceedings.neurips.cc/paper/2020/hash/1457c0d6bfcb4967418bfb8ac142f64a-Abstract.html} {Language models are few-shot learners}.
\newblock In \emph{Advances in Neural Information Processing Systems 33: Annual Conference on Neural Information Processing Systems 2020, NeurIPS 2020, December 6-12, 2020, virtual}.

\bibitem[{Chen et~al.(2023)Chen, Kim, Sriram, Durrett, and Choi}]{Chen2023ComplexCV}
Jifan Chen, Grace Kim, Aniruddh Sriram, Greg Durrett, and Eunsol Choi. 2023.
\newblock \href {https://api.semanticscholar.org/CorpusID:258822852} {Complex claim verification with evidence retrieved in the wild}.
\newblock \emph{ArXiv}, abs/2305.11859.

\bibitem[{Conneau et~al.(2020)Conneau, Khandelwal, Goyal, Chaudhary, Wenzek, Guzm{\'a}n, Grave, Ott, Zettlemoyer, and Stoyanov}]{conneau-etal-2020-unsupervised}
Alexis Conneau, Kartikay Khandelwal, Naman Goyal, Vishrav Chaudhary, Guillaume Wenzek, Francisco Guzm{\'a}n, Edouard Grave, Myle Ott, Luke Zettlemoyer, and Veselin Stoyanov. 2020.
\newblock \href {https://doi.org/10.18653/v1/2020.acl-main.747} {Unsupervised cross-lingual representation learning at scale}.
\newblock In \emph{Proceedings of the 58th Annual Meeting of the Association for Computational Linguistics}, pages 8440--8451, Online. Association for Computational Linguistics.

\bibitem[{Du et~al.(2022)Du, Qian, Liu, Ding, Qiu, Yang, and Tang}]{du2022glm}
Zhengxiao Du, Yujie Qian, Xiao Liu, Ming Ding, Jiezhong Qiu, Zhilin Yang, and Jie Tang. 2022.
\newblock \href {https://doi.org/10.18653/v1/2022.acl-long.26} {{GLM}: General language model pretraining with autoregressive blank infilling}.
\newblock In \emph{Proceedings of the 60th Annual Meeting of the Association for Computational Linguistics (Volume 1: Long Papers)}, pages 320--335, Dublin, Ireland. Association for Computational Linguistics.

\bibitem[{Fabbri et~al.(2019)Fabbri, Li, She, Li, and Radev}]{fabbri-etal-2019-multi}
Alexander Fabbri, Irene Li, Tianwei She, Suyi Li, and Dragomir Radev. 2019.
\newblock \href {https://doi.org/10.18653/v1/P19-1102} {Multi-news: A large-scale multi-document summarization dataset and abstractive hierarchical model}.
\newblock In \emph{Proceedings of the 57th Annual Meeting of the Association for Computational Linguistics}, pages 1074--1084, Florence, Italy. Association for Computational Linguistics.

\bibitem[{Fan et~al.(2019)Fan, Jernite, Perez, Grangier, Weston, and Auli}]{fan-etal-2019-eli5}
Angela Fan, Yacine Jernite, Ethan Perez, David Grangier, Jason Weston, and Michael Auli. 2019.
\newblock \href {https://doi.org/10.18653/v1/P19-1346} {{ELI}5: Long form question answering}.
\newblock In \emph{Proceedings of the 57th Annual Meeting of the Association for Computational Linguistics}, pages 3558--3567, Florence, Italy. Association for Computational Linguistics.

\bibitem[{Gao et~al.(2023{\natexlab{a}})Gao, Dai, Pasupat, Chen, Chaganty, Fan, Zhao, Lao, Lee, Juan, and Guu}]{gao-etal-2023-rarr}
Luyu Gao, Zhuyun Dai, Panupong Pasupat, Anthony Chen, Arun~Tejasvi Chaganty, Yicheng Fan, Vincent Zhao, Ni~Lao, Hongrae Lee, Da-Cheng Juan, and Kelvin Guu. 2023{\natexlab{a}}.
\newblock \href {https://doi.org/10.18653/v1/2023.acl-long.910} {{RARR}: Researching and revising what language models say, using language models}.
\newblock In \emph{Proceedings of the 61st Annual Meeting of the Association for Computational Linguistics (Volume 1: Long Papers)}, pages 16477--16508, Toronto, Canada. Association for Computational Linguistics.

\bibitem[{Gao et~al.(2023{\natexlab{b}})Gao, Yen, Yu, and Chen}]{Gao2023EnablingLL}
Tianyu Gao, Howard Yen, Jiatong Yu, and Danqi Chen. 2023{\natexlab{b}}.
\newblock \href {https://doi.org/10.18653/v1/2023.emnlp-main.398} {Enabling large language models to generate text with citations}.
\newblock In \emph{Proceedings of the 2023 Conference on Empirical Methods in Natural Language Processing}, pages 6465--6488, Singapore. Association for Computational Linguistics.

\bibitem[{Gehrmann et~al.(2023)Gehrmann, Clark, and Sellam}]{10.1613/jair.1.13715}
Sebastian Gehrmann, Elizabeth Clark, and Thibault Sellam. 2023.
\newblock \href {https://doi.org/10.1613/jair.1.13715} {Repairing the cracked foundation: A survey of obstacles in evaluation practices for generated text}.
\newblock \emph{J. Artif. Int. Res.}, 77.

\bibitem[{He et~al.(2023)He, Lin, Gong, Jin, Zhang, Lin, Jiao, Yiu, Duan, and Chen}]{He2023AnnoLLMML}
Xingwei He, Zheng-Wen Lin, Yeyun Gong, Alex Jin, Hang Zhang, Chen Lin, Jian Jiao, Siu~Ming Yiu, Nan Duan, and Weizhu Chen. 2023.
\newblock \href {https://arxiv.org/abs/2303.16854} {Annollm: Making large language models to be better crowdsourced annotators}.
\newblock \emph{ArXiv preprint}, abs/2303.16854.

\bibitem[{Holtzman et~al.(2020)Holtzman, Buys, Du, Forbes, and Choi}]{Holtzman2020The}
Ari Holtzman, Jan Buys, Li~Du, Maxwell Forbes, and Yejin Choi. 2020.
\newblock \href {https://openreview.net/forum?id=rygGQyrFvH} {The curious case of neural text degeneration}.
\newblock In \emph{International Conference on Learning Representations}.

\bibitem[{Huo et~al.(2023)Huo, Arabzadeh, and Clarke}]{10.1145/3624918.3625336}
Siqing Huo, Negar Arabzadeh, and Charles Clarke. 2023.
\newblock \href {https://doi.org/10.1145/3624918.3625336} {Retrieving supporting evidence for generative question answering}.
\newblock In \emph{Proceedings of the Annual International ACM SIGIR Conference on Research and Development in Information Retrieval in the Asia Pacific Region}, SIGIR-AP '23, page 11–20, New York, NY, USA. Association for Computing Machinery.

\bibitem[{Ji et~al.(2023)Ji, Lee, Frieske, Yu, Su, Xu, Ishii, Bang, Madotto, and Fung}]{10.1145/3571730}
Ziwei Ji, Nayeon Lee, Rita Frieske, Tiezheng Yu, Dan Su, Yan Xu, Etsuko Ishii, Ye~Jin Bang, Andrea Madotto, and Pascale Fung. 2023.
\newblock \href {https://doi.org/10.1145/3571730} {Survey of hallucination in natural language generation}.
\newblock \emph{ACM Comput. Surv.}, 55(12).

\bibitem[{Kamoi et~al.(2023)Kamoi, Goyal, Rodriguez, and Durrett}]{kamoi-etal-2023-wice}
Ryo Kamoi, Tanya Goyal, Juan Rodriguez, and Greg Durrett. 2023.
\newblock \href {https://aclanthology.org/2023.emnlp-main.470} {{W}i{CE}: Real-world entailment for claims in {W}ikipedia}.
\newblock In \emph{Proceedings of the 2023 Conference on Empirical Methods in Natural Language Processing}, pages 7561--7583, Singapore. Association for Computational Linguistics.

\bibitem[{Krishna et~al.(2021)Krishna, Roy, and Iyyer}]{krishna-etal-2021-hurdles}
Kalpesh Krishna, Aurko Roy, and Mohit Iyyer. 2021.
\newblock \href {https://doi.org/10.18653/v1/2021.naacl-main.393} {Hurdles to progress in long-form question answering}.
\newblock In \emph{Proceedings of the 2021 Conference of the North American Chapter of the Association for Computational Linguistics: Human Language Technologies}, pages 4940--4957, Online. Association for Computational Linguistics.

\bibitem[{Lewis et~al.(2021)Lewis, Stenetorp, and Riedel}]{lewis-etal-2021-question}
Patrick Lewis, Pontus Stenetorp, and Sebastian Riedel. 2021.
\newblock \href {https://doi.org/10.18653/v1/2021.eacl-main.86} {Question and answer test-train overlap in open-domain question answering datasets}.
\newblock In \emph{Proceedings of the 16th Conference of the European Chapter of the Association for Computational Linguistics: Main Volume}, pages 1000--1008, Online. Association for Computational Linguistics.

\bibitem[{Li et~al.(2023{\natexlab{a}})Li, Sun, Hu, Liu, Chen, Hu, Wu, and Zhang}]{li2023survey}
Dongfang Li, Zetian Sun, Xinshuo Hu, Zhenyu Liu, Ziyang Chen, Baotian Hu, Aiguo Wu, and Min Zhang. 2023{\natexlab{a}}.
\newblock \href {http://arxiv.org/abs/2311.03731} {A survey of large language models attribution}.

\bibitem[{Li and Li(2013)}]{li-li-2013-novel}
Jiwei Li and Sujian Li. 2013.
\newblock \href {https://doi.org/10.1162/tacl_a_00212} {A novel feature-based {B}ayesian model for query focused multi-document summarization}.
\newblock \emph{Transactions of the Association for Computational Linguistics}, 1:89--98.

\bibitem[{Li et~al.(2023{\natexlab{b}})Li, Zhu, Li, Yin, Sun, and Qiu}]{li2023llatrieval}
Xiaonan Li, Changtai Zhu, Linyang Li, Zhangyue Yin, Tianxiang Sun, and Xipeng Qiu. 2023{\natexlab{b}}.
\newblock \href {http://arxiv.org/abs/2311.07838} {Llatrieval: Llm-verified retrieval for verifiable generation}.

\bibitem[{Lin(2004)}]{lin-2004-rouge}
Chin-Yew Lin. 2004.
\newblock \href {https://aclanthology.org/W04-1013} {{ROUGE}: A package for automatic evaluation of summaries}.
\newblock In \emph{Text Summarization Branches Out}, pages 74--81, Barcelona, Spain. Association for Computational Linguistics.

\bibitem[{Liu et~al.(2023{\natexlab{a}})Liu, Zhang, and Liang}]{Liu2023EvaluatingVI}
Nelson~F. Liu, Tianyi Zhang, and Percy Liang. 2023{\natexlab{a}}.
\newblock \href {https://arxiv.org/abs/2304.09848} {Evaluating verifiability in generative search engines}.
\newblock \emph{ArXiv preprint}, abs/2304.09848.

\bibitem[{Liu et~al.(2023{\natexlab{b}})Liu, Lai, Yu, Xu, Zeng, Du, Zhang, Dong, and Tang}]{liu2023webglm}
Xiao Liu, Hanyu Lai, Hao Yu, Yifan Xu, Aohan Zeng, Zhengxiao Du, Peng Zhang, Yuxiao Dong, and Jie Tang. 2023{\natexlab{b}}.
\newblock \href {https://doi.org/10.1145/3580305.3599931} {Webglm: Towards an efficient web-enhanced question answering system with human preferences}.
\newblock In \emph{Proceedings of the 29th ACM SIGKDD Conference on Knowledge Discovery and Data Mining}, KDD '23, page 4549–4560, New York, NY, USA. Association for Computing Machinery.

\bibitem[{Liu et~al.(2023{\natexlab{c}})Liu, Fabbri, Liu, Zhao, Nan, Han, Han, Joty, Wu, Xiong, and Radev}]{liu-etal-2023-revisiting}
Yixin Liu, Alex Fabbri, Pengfei Liu, Yilun Zhao, Linyong Nan, Ruilin Han, Simeng Han, Shafiq Joty, Chien-Sheng Wu, Caiming Xiong, and Dragomir Radev. 2023{\natexlab{c}}.
\newblock \href {https://doi.org/10.18653/v1/2023.acl-long.228} {Revisiting the gold standard: Grounding summarization evaluation with robust human evaluation}.
\newblock In \emph{Proceedings of the 61st Annual Meeting of the Association for Computational Linguistics (Volume 1: Long Papers)}, pages 4140--4170, Toronto, Canada. Association for Computational Linguistics.

\bibitem[{Loshchilov and Hutter(2019)}]{loshchilov2018decoupled}
Ilya Loshchilov and Frank Hutter. 2019.
\newblock \href {https://openreview.net/forum?id=Bkg6RiCqY7} {Decoupled weight decay regularization}.
\newblock In \emph{International Conference on Learning Representations}.

\bibitem[{Micikevicius et~al.(2018)Micikevicius, Narang, Alben, Diamos, Elsen, Garcia, Ginsburg, Houston, Kuchaiev, Venkatesh, and Wu}]{micikevicius2018mixed}
Paulius Micikevicius, Sharan Narang, Jonah Alben, Gregory Diamos, Erich Elsen, David Garcia, Boris Ginsburg, Michael Houston, Oleksii Kuchaiev, Ganesh Venkatesh, and Hao Wu. 2018.
\newblock \href {http://arxiv.org/abs/1710.03740} {Mixed precision training}.

\bibitem[{Min et~al.(2023)Min, Krishna, Lyu, Lewis, Yih, Koh, Iyyer, Zettlemoyer, and Hajishirzi}]{min-etal-2023-factscore}
Sewon Min, Kalpesh Krishna, Xinxi Lyu, Mike Lewis, Wen-tau Yih, Pang Koh, Mohit Iyyer, Luke Zettlemoyer, and Hannaneh Hajishirzi. 2023.
\newblock \href {https://doi.org/10.18653/v1/2023.emnlp-main.741} {{FA}ct{S}core: Fine-grained atomic evaluation of factual precision in long form text generation}.
\newblock In \emph{Proceedings of the 2023 Conference on Empirical Methods in Natural Language Processing}, pages 12076--12100, Singapore. Association for Computational Linguistics.

\bibitem[{Nakano et~al.(2021)Nakano, Hilton, Balaji, Wu, Long, Kim, Hesse, Jain, Kosaraju, Saunders, Jiang, Cobbe, Eloundou, Krueger, Button, Knight, Chess, and Schulman}]{Nakano2021WebGPTBQ}
Reiichiro Nakano, Jacob Hilton, S.~Arun Balaji, Jeff Wu, Ouyang Long, Christina Kim, Christopher Hesse, Shantanu Jain, Vineet Kosaraju, William Saunders, Xu~Jiang, Karl Cobbe, Tyna Eloundou, Gretchen Krueger, Kevin Button, Matthew Knight, Benjamin Chess, and John Schulman. 2021.
\newblock \href {https://arxiv.org/abs/2112.09332} {Webgpt: Browser-assisted question-answering with human feedback}.
\newblock \emph{ArXiv preprint}, abs/2112.09332.

\bibitem[{Nenkova and Passonneau(2004)}]{nenkova-passonneau-2004-evaluating}
Ani Nenkova and Rebecca Passonneau. 2004.
\newblock \href {https://aclanthology.org/N04-1019} {Evaluating content selection in summarization: The pyramid method}.
\newblock In \emph{Proceedings of the Human Language Technology Conference of the North {A}merican Chapter of the Association for Computational Linguistics: {HLT}-{NAACL} 2004}, pages 145--152, Boston, Massachusetts, USA. Association for Computational Linguistics.

\bibitem[{Ouyang et~al.(2022)Ouyang, Wu, Jiang, Almeida, Wainwright, Mishkin, Zhang, Agarwal, Slama, Ray et~al.}]{ouyang2022training}
Long Ouyang, Jeffrey Wu, Xu~Jiang, Diogo Almeida, Carroll Wainwright, Pamela Mishkin, Chong Zhang, Sandhini Agarwal, Katarina Slama, Alex Ray, et~al. 2022.
\newblock Training language models to follow instructions with human feedback.
\newblock \emph{Advances in Neural Information Processing Systems}, 35:27730--27744.

\bibitem[{Passonneau(2009)}]{Passonneau2009FormalAF}
R.~Passonneau. 2009.
\newblock \href {https://api.semanticscholar.org/CorpusID:17543253} {Formal and functional assessment of the pyramid method for summary content evaluation*}.
\newblock \emph{Natural Language Engineering}, 16:107 -- 131.

\bibitem[{Post(2018)}]{post-2018-call}
Matt Post. 2018.
\newblock \href {https://www.aclweb.org/anthology/W18-6319} {A call for clarity in reporting {BLEU} scores}.
\newblock In \emph{Proceedings of the Third Conference on Machine Translation: Research Papers}, pages 186--191, Belgium, Brussels. Association for Computational Linguistics.

\bibitem[{Qin et~al.(2023)Qin, Cai, Jin, Yan, Liang, Zhu, Lin, Han, Ding, Wang, Xie, Qi, Liu, Sun, and Zhou}]{qin2023webcpm}
Yujia Qin, Zihan Cai, Dian Jin, Lan Yan, Shihao Liang, Kunlun Zhu, Yankai Lin, Xu~Han, Ning Ding, Huadong Wang, Ruobing Xie, Fanchao Qi, Zhiyuan Liu, Maosong Sun, and Jie Zhou. 2023.
\newblock \href {https://doi.org/10.18653/v1/2023.acl-long.499} {{W}eb{CPM}: Interactive web search for {C}hinese long-form question answering}.
\newblock In \emph{Proceedings of the 61st Annual Meeting of the Association for Computational Linguistics (Volume 1: Long Papers)}, pages 8968--8988, Toronto, Canada. Association for Computational Linguistics.

\bibitem[{Rajbhandari et~al.(2020)Rajbhandari, Rasley, Ruwase, and He}]{Rajbhandari2020zero}
Samyam Rajbhandari, Jeff Rasley, Olatunji Ruwase, and Yuxiong He. 2020.
\newblock Zero: Memory optimizations toward training trillion parameter models.
\newblock In \emph{Proceedings of the International Conference for High Performance Computing, Networking, Storage and Analysis}, SC '20. IEEE Press.

\bibitem[{Rashkin et~al.(2023)Rashkin, Nikolaev, Lamm, Aroyo, Collins, Das, Petrov, Tomar, Turc, and Reitter}]{Rashkin2023attr}
Hannah Rashkin, Vitaly Nikolaev, Matthew Lamm, Lora Aroyo, Michael Collins, Dipanjan Das, Slav Petrov, Gaurav~Singh Tomar, Iulia Turc, and David Reitter. 2023.
\newblock \href {https://doi.org/10.1162/coli_a_00486} {{Measuring Attribution in Natural Language Generation Models}}.
\newblock \emph{Computational Linguistics}, pages 1--64.

\bibitem[{Roy and Kundu(2023)}]{qfmdsreview}
Prasenjeet Roy and Suman Kundu. 2023.
\newblock \href {https://doi.org/10.1145/3597299} {Review on query-focused multi-document summarization (qmds) with comparative analysis}.
\newblock \emph{ACM Comput. Surv.}, 56(1).

\bibitem[{Shapira et~al.(2019)Shapira, Gabay, Gao, Ronen, Pasunuru, Bansal, Amsterdamer, and Dagan}]{shapira-etal-2019-crowdsourcing}
Ori Shapira, David Gabay, Yang Gao, Hadar Ronen, Ramakanth Pasunuru, Mohit Bansal, Yael Amsterdamer, and Ido Dagan. 2019.
\newblock \href {https://doi.org/10.18653/v1/N19-1072} {Crowdsourcing lightweight pyramids for manual summary evaluation}.
\newblock In \emph{Proceedings of the 2019 Conference of the North {A}merican Chapter of the Association for Computational Linguistics: Human Language Technologies, Volume 1 (Long and Short Papers)}, pages 682--687, Minneapolis, Minnesota. Association for Computational Linguistics.

\bibitem[{Shaw et~al.(2018)Shaw, Uszkoreit, and Vaswani}]{shaw-etal-2018-self}
Peter Shaw, Jakob Uszkoreit, and Ashish Vaswani. 2018.
\newblock \href {https://doi.org/10.18653/v1/N18-2074} {Self-attention with relative position representations}.
\newblock In \emph{Proceedings of the 2018 Conference of the North {A}merican Chapter of the Association for Computational Linguistics: Human Language Technologies, Volume 2 (Short Papers)}, pages 464--468, New Orleans, Louisiana. Association for Computational Linguistics.

\bibitem[{Shoeybi et~al.(2019)Shoeybi, Patwary, Puri, LeGresley, Casper, and Catanzaro}]{Shoeybi2019MegatronLMTM}
Mohammad Shoeybi, Mostofa Patwary, Raul Puri, Patrick LeGresley, Jared Casper, and Bryan Catanzaro. 2019.
\newblock \href {https://arxiv.org/abs/1909.08053} {Megatron-lm: Training multi-billion parameter language models using model parallelism}.
\newblock \emph{ArXiv preprint}, abs/1909.08053.

\bibitem[{Stelmakh et~al.(2022)Stelmakh, Luan, Dhingra, and Chang}]{stelmakh-etal-2022-asqa}
Ivan Stelmakh, Yi~Luan, Bhuwan Dhingra, and Ming-Wei Chang. 2022.
\newblock \href {https://aclanthology.org/2022.emnlp-main.566} {{ASQA}: Factoid questions meet long-form answers}.
\newblock In \emph{Proceedings of the 2022 Conference on Empirical Methods in Natural Language Processing}, pages 8273--8288, Abu Dhabi, United Arab Emirates. Association for Computational Linguistics.

\bibitem[{Tombros and Sanderson(1998)}]{tombros1998advantages}
Anastasios Tombros and Mark Sanderson. 1998.
\newblock Advantages of query biased summaries in information retrieval.
\newblock In \emph{Proceedings of the 21st annual international ACM SIGIR conference on Research and development in information retrieval}, pages 2--10.

\bibitem[{Weller et~al.(2023)Weller, Marone, Weir, Lawrie, Khashabi, and Durme}]{weller2023according}
Orion Weller, Marc Marone, Nathaniel Weir, Dawn Lawrie, Daniel Khashabi, and Benjamin~Van Durme. 2023.
\newblock \href {http://arxiv.org/abs/2305.13252} {"according to ..." prompting language models improves quoting from pre-training data}.

\bibitem[{Xu et~al.(2023)Xu, Song, Iyyer, and Choi}]{xu-etal-2023-critical}
Fangyuan Xu, Yixiao Song, Mohit Iyyer, and Eunsol Choi. 2023.
\newblock \href {https://doi.org/10.18653/v1/2023.acl-long.181} {A critical evaluation of evaluations for long-form question answering}.
\newblock In \emph{Proceedings of the 61st Annual Meeting of the Association for Computational Linguistics (Volume 1: Long Papers)}, pages 3225--3245, Toronto, Canada. Association for Computational Linguistics.

\bibitem[{Xue et~al.(2021)Xue, Constant, Roberts, Kale, Al-Rfou, Siddhant, Barua, and Raffel}]{xue-etal-2021-mt5}
Linting Xue, Noah Constant, Adam Roberts, Mihir Kale, Rami Al-Rfou, Aditya Siddhant, Aditya Barua, and Colin Raffel. 2021.
\newblock \href {https://doi.org/10.18653/v1/2021.naacl-main.41} {m{T}5: A massively multilingual pre-trained text-to-text transformer}.
\newblock In \emph{Proceedings of the 2021 Conference of the North American Chapter of the Association for Computational Linguistics: Human Language Technologies}, pages 483--498, Online. Association for Computational Linguistics.

\bibitem[{Yang et~al.(2023)Yang, Xiao, Wang, Zhang, Bian, Yin, Lv, Pan, Wang, Yan, Yang, Deng, Wang, Liu, Ai, Dong, Zhao, Xu, Sun, Zhang, Liu, Ji, Xie, Dai, Fang, Su, Song, Liu, Ru, Ma, Wang, Liu, Lin, Nie, Guo, Sun, Tao, Li, Li, Cheng, Chen, Zeng, Wang, Chen, Men, Yu, Pan, Shen, Wang, Li, Jiang, Gao, Zhang, Zhou, and Wu}]{Yang2023Baichuan2O}
Ai~Ming Yang, Bin Xiao, Bingning Wang, Borong Zhang, Ce~Bian, Chao Yin, Chenxu Lv, Da~Pan, Dian Wang, Dong Yan, Fan Yang, Fei Deng, Feng Wang, Feng Liu, Guangwei Ai, Guosheng Dong, Hai Zhao, Hang Xu, Hao Sun, Hongda Zhang, Hui Liu, Jiaming Ji, Jian Xie, Juntao Dai, Kuncheng Fang, Lei Su, Liang Song, Lifeng Liu, Liyun Ru, Luyao Ma, Mang Wang, Mickel Liu, MingAn Lin, Nuolan Nie, Pei Guo, Ruiyang Sun, Zhang Tao, Tianpeng Li, Tianyu Li, Wei Cheng, Weipeng Chen, Xiangrong Zeng, Xiaochuan Wang, Xiaoxi Chen, Xin Men, Xin Yu, Xuehai Pan, Yan-Bin Shen, Yiding Wang, Yiyu Li, Youxin Jiang, Yuchen Gao, Yupeng Zhang, Zenan Zhou, and Zhiying Wu. 2023.
\newblock \href {https://api.semanticscholar.org/CorpusID:261951743} {Baichuan 2: Open large-scale language models}.
\newblock \emph{ArXiv}, abs/2309.10305.

\bibitem[{Yue et~al.(2023)Yue, Wang, Zhang, Chen, Su, and Sun}]{Yue2023AutomaticAttr}
Xiang Yue, Boshi Wang, Kai Zhang, Ziru Chen, Yu~Su, and Huan Sun. 2023.
\newblock \href {https://arxiv.org/abs/2305.06311} {Automatic evaluation of attribution by large language models}.
\newblock \emph{ArXiv preprint}, abs/2305.06311.

\bibitem[{Zhang et~al.(2022)Zhang, Gan, Wang, Zhang, Zhang, Yang, Gao, Wu, Dong, He, Zhuo, Yang, Huang, Li, Wu, Lu, Zhu, Chen, Han, Pan, Wang, Wang, Wu, Zeng, and Chen}]{fengshenbang}
Jiaxing Zhang, Ruyi Gan, Junjie Wang, Yuxiang Zhang, Lin Zhang, Ping Yang, Xinyu Gao, Ziwei Wu, Xiaoqun Dong, Junqing He, Jianheng Zhuo, Qi~Yang, Yongfeng Huang, Xiayu Li, Yanghan Wu, Junyu Lu, Xinyu Zhu, Weifeng Chen, Ting Han, Kunhao Pan, Rui Wang, Hao Wang, Xiaojun Wu, Zhongshen Zeng, and Chongpei Chen. 2022.
\newblock \href {https://arxiv.org/abs/2209.02970} {Fengshenbang 1.0: Being the foundation of chinese cognitive intelligence}.
\newblock \emph{ArXiv preprint}, abs/2209.02970.

\bibitem[{Zhang et~al.(2023{\natexlab{a}})Zhang, Dong, Li, Zhang, Sun, Wang, Li, Hu, Zhang, Wu, and Wang}]{zhang2023instruction}
Shengyu Zhang, Linfeng Dong, Xiaoya Li, Sen Zhang, Xiaofei Sun, Shuhe Wang, Jiwei Li, Runyi Hu, Tianwei Zhang, Fei Wu, and Guoyin Wang. 2023{\natexlab{a}}.
\newblock \href {http://arxiv.org/abs/2308.10792} {Instruction tuning for large language models: A survey}.

\bibitem[{Zhang and Bansal(2021)}]{zhang-bansal-2021-finding}
Shiyue Zhang and Mohit Bansal. 2021.
\newblock \href {https://doi.org/10.18653/v1/2021.emnlp-main.531} {Finding a balanced degree of automation for summary evaluation}.
\newblock In \emph{Proceedings of the 2021 Conference on Empirical Methods in Natural Language Processing}, pages 6617--6632, Online and Punta Cana, Dominican Republic. Association for Computational Linguistics.

\bibitem[{Zhang et~al.(2023{\natexlab{b}})Zhang, Ladhak, Durmus, Liang, McKeown, and Hashimoto}]{Zhang2023BenchmarkingLL}
Tianyi Zhang, Faisal Ladhak, Esin Durmus, Percy Liang, Kathleen McKeown, and Tatsunori Hashimoto. 2023{\natexlab{b}}.
\newblock \href {https://arxiv.org/abs/2301.13848} {Benchmarking large language models for news summarization}.
\newblock \emph{ArXiv preprint}, abs/2301.13848.

\bibitem[{Zhang et~al.(2023{\natexlab{c}})Zhang, Li, Cui, Cai, Liu, Fu, Huang, Zhao, Zhang, Chen, Wang, Luu, Bi, Shi, and Shi}]{zhang2023sirens}
Yue Zhang, Yafu Li, Leyang Cui, Deng Cai, Lemao Liu, Tingchen Fu, Xinting Huang, Enbo Zhao, Yu~Zhang, Yulong Chen, Longyue Wang, Anh~Tuan Luu, Wei Bi, Freda Shi, and Shuming Shi. 2023{\natexlab{c}}.
\newblock \href {http://arxiv.org/abs/2309.01219} {Siren's song in the ai ocean: A survey on hallucination in large language models}.

\bibitem[{Zhao et~al.(2023)Zhao, Zhou, Li, Tang, Wang, Hou, Min, Zhang, Zhang, Dong, Du, Yang, Chen, Chen, Jiang, Ren, Li, Tang, Liu, Liu, Nie, and rong Wen}]{Zhao2023ASO}
Wayne~Xin Zhao, Kun Zhou, Junyi Li, Tianyi Tang, Xiaolei Wang, Yupeng Hou, Yingqian Min, Beichen Zhang, Junjie Zhang, Zican Dong, Yifan Du, Chen Yang, Yushuo Chen, Z.~Chen, Jinhao Jiang, Ruiyang Ren, Yifan Li, Xinyu Tang, Zikang Liu, Peiyu Liu, Jianyun Nie, and Ji~rong Wen. 2023.
\newblock \href {https://arxiv.org/abs/2303.18223} {A survey of large language models}.
\newblock \emph{ArXiv preprint}, abs/2303.18223.

\bibitem[{Zhou et~al.(2023)Zhou, Lu, Mishra, Brahma, Basu, Luan, Zhou, and Hou}]{zhou2023instructionfollowing}
Jeffrey Zhou, Tianjian Lu, Swaroop Mishra, Siddhartha Brahma, Sujoy Basu, Yi~Luan, Denny Zhou, and Le~Hou. 2023.
\newblock \href {http://arxiv.org/abs/2311.07911} {Instruction-following evaluation for large language models}.

\bibitem[{Zhu et~al.(2018)Zhu, Lu, Zheng, Guo, Zhang, Wang, and Yu}]{zhu18Texygen}
Yaoming Zhu, Sidi Lu, Lei Zheng, Jiaxian Guo, Weinan Zhang, Jun Wang, and Yong Yu. 2018.
\newblock \href {https://doi.org/10.1145/3209978.3210080} {Texygen: A benchmarking platform for text generation models}.
\newblock In \emph{The 41st International ACM SIGIR Conference on Research \& Development in Information Retrieval}, SIGIR '18, page 1097–1100, New York, NY, USA. Association for Computing Machinery.

\end{thebibliography}

\appendix

\section{The WebCiteS Dataset}
\label{sec:apdx_dataset}

\begin{figure*}
\centering
    \includegraphics[width=\linewidth]{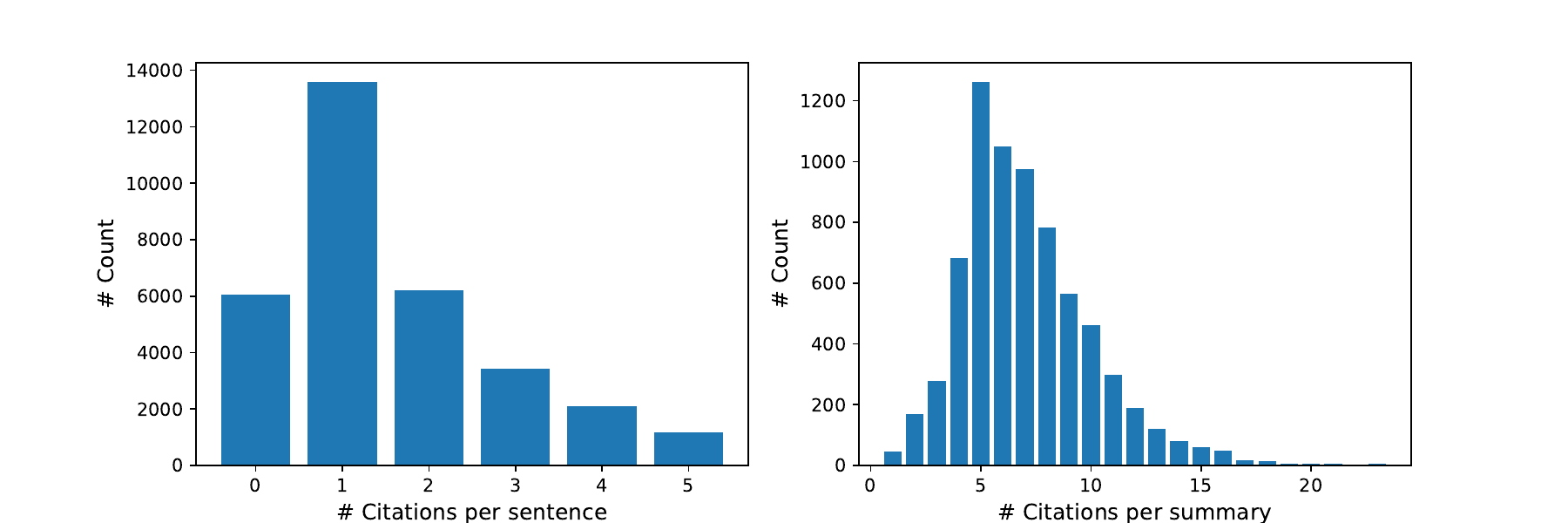}
\caption{The distribution of the number of citations per sentence and summary in WebCiteS.}
\label{fig:citation_dist}
\end{figure*}

\begin{figure*}
\centering
\vspace{-1em}
\includegraphics[width=\linewidth]{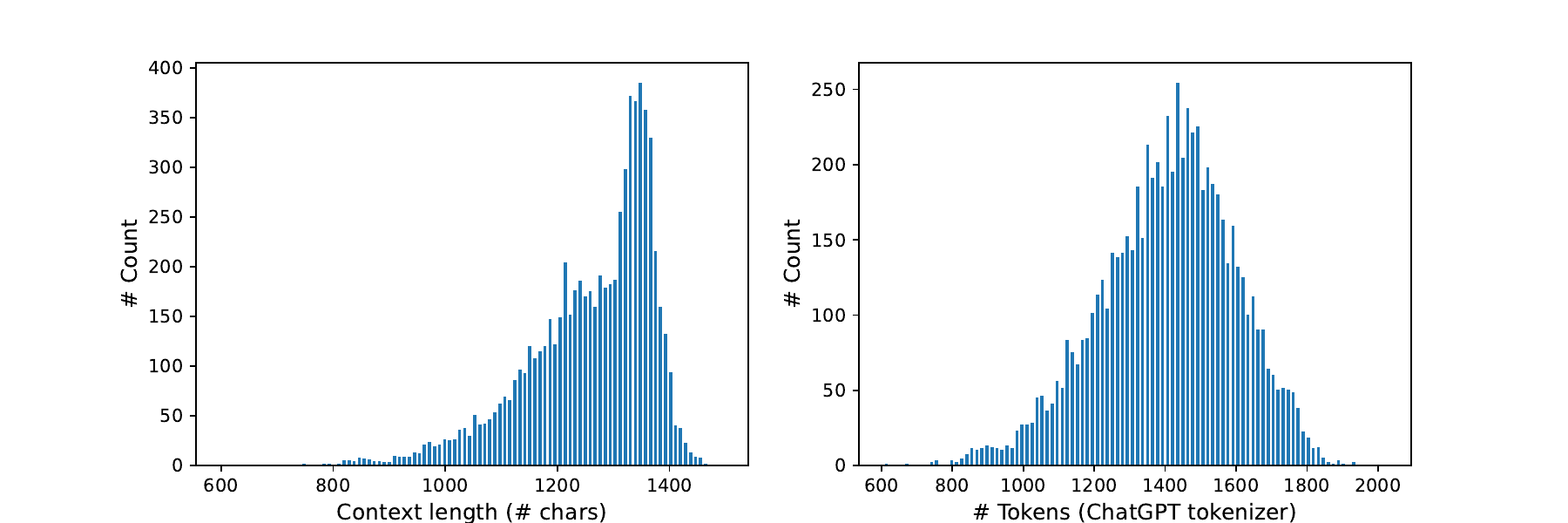}
\caption{The distribution of the input context length and the number of input tokens in the default setting, where each sample consists of five snippets of web pages as documents. We use the tokenizer of gpt-3.5-turbo.}
\label{fig:snippet_len_dist}
\end{figure*}

\begin{figure*}
\centering
\vspace{-1em}
\includegraphics[width=\linewidth]{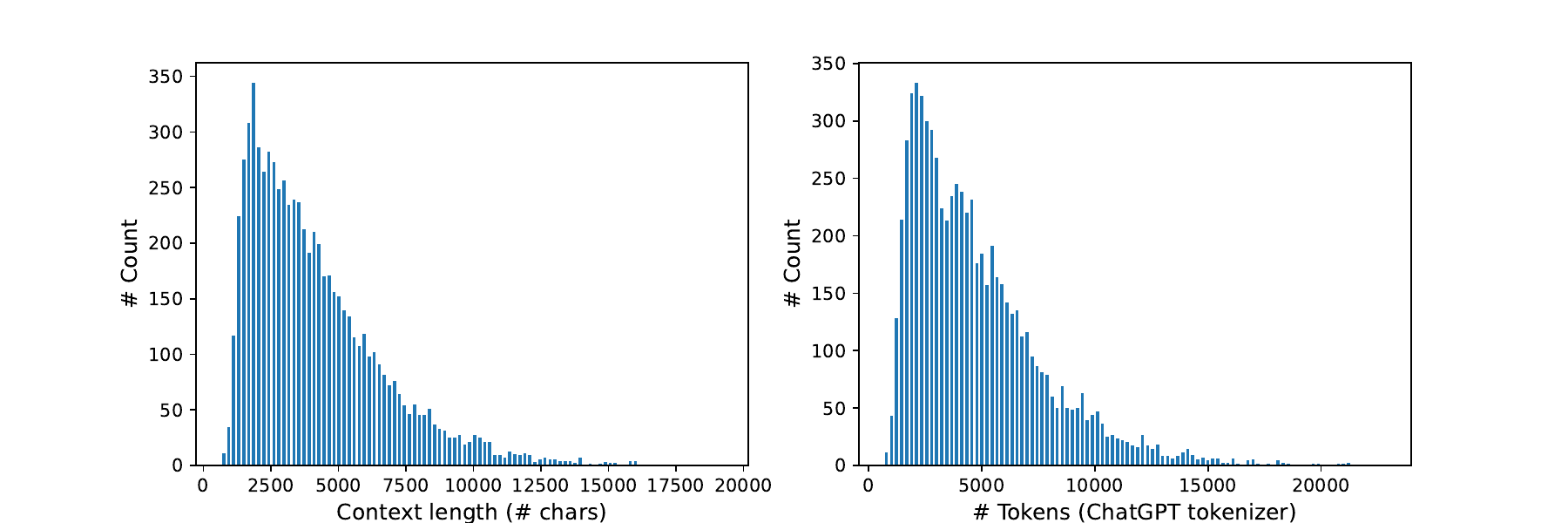}
\caption{The distribution of the input context length and the number of input tokens in the long-context setting, where we chunk the full content of web pages into documents with a maximum length of 512 characters. We use the tokenizer of gpt-3.5-turbo.}
\label{fig:fulldoc_len_dist}
\end{figure*}

Table~\ref{tab:fsp_prompt} presents an annotated example in WebCiteS. Table~\ref{tab:domain_dist} displays the domain distribution of the user queries. Figure~\ref{fig:citation_dist} displays the distributions of citation numbers. The distribution of context length for using snippets or full content as documents are shown in Figure~\ref{fig:snippet_len_dist} and Figure~\ref{fig:fulldoc_len_dist} respectively.

\begin{table*}[]
\centering
\vspace{-1em}
\small
\begin{tabular}{@{}rlc|rlc@{}}
\toprule
\multicolumn{2}{c}{\textbf{Domain}}      & \textbf{Count} & \multicolumn{2}{c}{\textbf{Domain}}        & \textbf{Count} \\ \midrule
生活知识 & Daily Life Knowledge     & 1370  & 金融   & Finance                    & 351   \\
教育培训 & Education and Training   & 1032  & 房产装修 & Real Estate and Decoration & 256   \\
政策法规 & Policies and Regulations & 726   & 生产制造 & Manufacturing              & 221   \\
商品   & Commodities              & 682   & 游戏娱乐 & Gaming and Entertainment   & 182   \\
其他   & Others                   & 508   & 交通出行 & Transportation             & 158   \\
机动车  & Vehicles                 & 430   & 旅游   & Travel                     & 146   \\
信息技术 & Information Technology   & 424   & 母婴育儿 & Maternity and Childcare    & 135   \\
动植物  & Flora and Fauna          & 398   & 民俗文化 & Folk Culture               & 111   \\ \bottomrule
\end{tabular}
\caption{The domain distribution of the user queries in WebCiteS, covering a broad range of real-world scenarios.}
\label{tab:domain_dist}
\end{table*}

\subsection{More Details of Data Annotation}
\label{subsec:apdx_annotation}

\paragraph{Sample selection.}
After 40,000 raw queries were gathered,  we adopted a rule-based system to remove common trivial queries. We also filter out queries seeking health and medicine advice, since these scenarios are of high risks and hard to judge without professional expertise. Through this process, we obtain 18,500 filtered queries.
To ensure the quality of the data, the remaining samples were further filtered by human annotators. Specifically, a sample would be discarded if it matched the following scenarios:
\begin{enumerate}
    \item If the query was too trivial and did not need long-form answers, or if it was seeking creative inspirations which did not need to be supported by evidence.
    \item If the query did not express its demand clearly and was hard to understand.
    \item If the query and documents contained inappropriate content, such as personal information, prejudice or bias against specific groups, and controversial topics.
    \item If the query could not be answered by the retrieved documents, or the reliability of certain documents was highly questionable to the annotators.
\end{enumerate}

\paragraph{Stage~1: manual screening and information extraction.}
We developed annotation software which allowed annotators to highlight clause-level segments containing useful information. The manual filtering of invalid samples based on the above criteria also took place in this stage.

\paragraph{Stage~2: LLM-based candidate summary generation.}
In the early annotation phase, We employed ChatGPT to summarize the information extracted from each sample. As our dataset grew, after accumulating 1.2k samples, we additionally fine-tuned an open-source model, ChatGLM2-6B, to provide an extra candidate summary for each sample. We upgraded this model iteratively with the influx of new annotations. Instead of generating multiple outputs by ChatGPT, the motivation for fine-tuning an extra model is to increase the diversity of the candidates. 

\paragraph{Stage~3: manual refinement and citation annotation.}
We outline a streamlined refinement process as follows: first, annotators were instructed to examine each sentence in the chosen summary, removing unimportant or redundant content. The importance of content was based on the extracted information from the first stage.
After that, annotators would identify the verification-worthy sentences in the summary, compare them with all documents with highlighted extraction, and cite all supporting ones. They would also rectify any hallucinations or groundedness errors in the sentences detected during citation annotation. After adding citations, they further ensured all the extracted information was referenced by the summary. If any important information was missing, they would either expand existing sentences or craft new ones to supplement the information, thereby enriching the comprehensiveness of the summary. Finally, annotators inspected the entire summary once again and refined its writing to improve coherence. They were also encouraged to add an introductory sentence to the beginning of the summary to enhance its readability.

\section{Evaluation metrics}
\label{sec:apdx_metrics}
We bring more details and discussions on the evaluation metrics.

\subsection{Metrics of Summarization Utility}
\label{subsec:apdx_summ_metrics}
\paragraph{Length.}
It is measured in the number of characters. We remove all citations in the summary before computing length. 

\paragraph{Self-BLEU.}
We compute Self-BLEU based on BLEU-4. Our implementation of self-BLEU is based on the sacreBLEU library~\citep{post-2018-call}.

\paragraph{Claim F$_1$.}
Many prior works in summarization evaluation follow the Pyramid protocol~\citep{nenkova-passonneau-2004-evaluating} to decompose the reference summaries into summary content units~\citep{Passonneau2009FormalAF, shapira-etal-2019-crowdsourcing, zhang-bansal-2021-finding} or atomic content units~\citep{liu-etal-2023-revisiting}, and measure how many units are covered in the system summaries (i.e., recall-based metrics). One advantage of recall-based metrics is that only reference summaries need to be decomposed since early works mostly rely on human efforts for sentence decomposition. On the other hand, recent studies also investigate the use of proprietary LLMs~\cite{Gao2023EnablingLL,min-etal-2023-factscore,kamoi-etal-2023-wice}. For example, \citet{Gao2023EnablingLL} use InstructGPT~\cite{ouyang2022training} to generate three sub-claims for each reference answer and compute claim recall to measure the correctness of model generations. However, the cost and rate limits of proprietary LLMs still hinder the scalability. In this work, we train a tailored claim-split model that enables us to calculate both claim precision and recall by extracting all sub-claims from both the system summaries and reference summaries with minimum cost.
Moreover, we do not use traditional automatic metrics such as ROUGE~\citep{lin-2004-rouge} since their limitations in evaluating LLM generations have been discussed in recent works.~\citep{Zhang2023BenchmarkingLL,gao-etal-2023-rarr,xu-etal-2023-critical}.

\subsection{Metrics of Attribution}
\label{subsec:apdx_attr_metrics}

\paragraph{Granularity.}
We evaluate all metrics of attribution at the sentence level, primarily because the citations in the WebCiteS are annotated at the sentence level, as more fine-grained annotation would require annotators to manually extract sub-claims from a sentence which bring extra cost. However, future works could extend these metrics to the sub-claim level depending on their needs.


\paragraph{Citation precision.}
\citet{Liu2023EvaluatingVI,Gao2023EnablingLL} compute citation precision by calculating the fraction of accurate citations within the whole generation. The major differences in our approach are:
\begin{enumerate}[leftmargin=12pt]
    \item We try to look for $C^{i*}_{\text{Pred}}$ if $C^{i}_{\text{Pred}}$ is empty to avoid unnecessary penalization which leads to the undervaluation of model performance.
    \item We compute citation precision at the sentence level and average them for the whole summary, while \citet{Liu2023EvaluatingVI,Gao2023EnablingLL} compute this metric directly at the response-level.
\end{enumerate}

\paragraph{Citation recall.}
\citet{Liu2023EvaluatingVI,Gao2023EnablingLL} define citation recall as the fraction of sentences being fully supported by their citations. This is equivalent to the AIS score~\citep{Bohnet2022AttributedQA,Rashkin2023attr,gao-etal-2023-rarr} by taking citations as the \textit{identified sources}. In contrast, our definition of citation recall is consistent with citation precision by calculating the fraction of citations, which is aligned with the naming of the metric.

\begin{table}[t]
\centering
\small
\setlength\tabcolsep{4pt}
\renewcommand{\arraystretch}{1.1}
\begin{tabular}{@{}cccc@{}}
\toprule
                              & \multicolumn{3}{c}{\textbf{Citation Mask}}        \\
\multicolumn{1}{l}{\textbf{}} & \textbf{Default} & \textbf{Auto} & \textbf{Human} \\ \midrule
XLM-RoBERTa-Large-XNLI     & 75.1                 & 78.0              & 78.4               \\
ELS-RoBERTa-Large-NLI     & 74.2                 & 76.2              & 77.6               \\
ELS-MBERT-1.3B-NLI        & 76.0                 & 79.1              & 79.3               \\
mT5-Large-XNLI             & \textbf{78.6}        & \textbf{82.3}     & \textbf{82.6}      \\ \bottomrule
\end{tabular}
\caption{Performance of citation prediction using different NLI models under three citation mask settings.}
\label{tab:apdx_nli_evaluation}
\end{table}

\section{Experiments on NLI Model}
\label{sec:apdx_nli}
We evaluate the performance of different NLI models via a citation prediction task on the test set of WebCiteS: for each sentence in the summary and each given document, we use the NLI model to classify whether the sentence should cite the document, and calculate its accuracy by taking human citations as ground truth. Only sentences with citation mask $m_i=1$ are considered. We adopt three citation mask settings: \textit{default}, \textit{auto}, and \textit{human}, similar to the experiments in Section~\ref{subsec:test_auto_eval}. We select the following NLI models for evaluation: 
(1) XLM-RoBERTa-Large-XNLI, an XLM-RoBERTa model~\citep{conneau-etal-2020-unsupervised} fine-tuned on multilingual NLI datasets.\footnote{\url{https://huggingface.co/joeddav/xlm-roberta-large-xnli}}
(2) ELS-RoBERTa-Large-NLI, a Chinese RoBERTa model fine-tuned on several NLI datasets~\citep{fengshenbang}.\footnote{\url{https://huggingface.co/IDEA-CCNL/Erlangshen-Roberta-330M-NLI}}
(3) ELS-MBERT-1.3B-NLI, a Chinese model based on the MegatronBERT architecture~\citep{Shoeybi2019MegatronLMTM}, fine-tuned on several NLI datasets~\citep{fengshenbang}.\footnote{\url{https://huggingface.co/IDEA-CCNL/Erlangshen-MegatronBert-1.3B-NLI}}, 
(4) mT5-Large-XNLI, an mT5 model~\citep{xue-etal-2021-mt5} fine-tuned on multilingual NLI datasets.\footnote{\url{https://huggingface.co/alan-turing-institute/mt5-large-finetuned-mnli-xtreme-xnli}}

The results in Table~\ref{tab:apdx_nli_evaluation} show that the mT5 model achieves the highest accuracy in predicting citations. Moreover, using the default citation mask (i.e., set $m_i=1$ for all sentences) lowers accuracy across all models, underscoring the necessity of identifying if a sentence is verification-worthy. Besides, we find that results under auto citation mask and human citation mask are notably similar. This validates the effectiveness of our citation mask prediction method proposed in Section~\ref{subsec:attribution_metrics}.

\begin{table*}[t]
\centering
\small
\renewcommand{\arraystretch}{1.1}
\begin{tabular}{@{}ccccccccccccc@{}}
\toprule
 &
   &
   &
   &
   &
  \multicolumn{3}{c}{\textbf{Claim}} &
  \multicolumn{3}{c}{\textbf{Citation}} &
   &
   \\
\multirow{-2}{*}{\textbf{\begin{tabular}[c]{@{}c@{}}Source\\ Type\end{tabular}}} &
  \multirow{-2}{*}{\textbf{\begin{tabular}[c]{@{}c@{}}Max Doc \\ Length\end{tabular}}} &
  \multirow{-2}{*}{\textbf{\# Docs}} &
  \multirow{-2}{*}{\textbf{Len.}} &
  \multirow{-2}{*}{\textbf{\begin{tabular}[c]{@{}c@{}}Self-\\ Bleu ↓\end{tabular}}} &
  \textbf{Prec.} &
  \textbf{Rec.} &
  \textbf{F$_1$} &
  \textbf{Prec.} &
  \textbf{Rec.} &
  \textbf{F$_1$} &
  \multirow{-2}{*}{\textbf{AIS}} &
  \multirow{-2}{*}{\textbf{ACS}} \\ \midrule
\multicolumn{13}{c}{\cellcolor[HTML]{EFEFEF}\textbf{ChatGPT}} \\
Snippet &
  250 &
  5 &
  223 &
  12.2 &
  53.5 &
  \textbf{56.5} &
  \textbf{54.9} &
  71.2 &
  64.8 &
  67.8 &
  75.1 &
  84.7 \\
Full Content &
  512 &
  11 &
  238 &
  10.5 &
  42.9 &
  49.3 &
  45.9 &
  63.6 &
  53.6 &
  58.2 &
  72.2 &
  88.9 \\
Full Content &
  256 &
  20 &
  245 &
  10.6 &
  41.2 &
  49.3 &
  44.9 &
  58.2 &
  45.9 &
  51.3 &
  65.2 &
  86.5 \\
\multicolumn{13}{c}{\cellcolor[HTML]{EFEFEF}\textbf{ChatGLM3-6B (SFT)}} \\
Snippet &
  250 &
  5 &
  163 &
  \textbf{8.9} &
  \textbf{55.7} &
  49.3 &
  52.3 &
  \textbf{78.5} &
  \textbf{73.8} &
  \textbf{76.1} &
  \textbf{81.3} &
  \textbf{89.4} \\
Full Content &
  512 &
  11 &
  170 &
  \textbf{8.9} &
  45.6 &
  40.8 &
  43.1 &
  61.7 &
  44.7 &
  51.9 &
  65.8 &
  88.3 \\
Full Content &
  256 &
  20 &
  173 &
  \textbf{8.9} &
  44.5 &
  39.4 &
  41.8 &
  53.1 &
  35.9 &
  42.9 &
  56.3 &
  84.9 \\
\multicolumn{13}{c}{\cellcolor[HTML]{EFEFEF}\textbf{Baichuan2-7B (SFT)}} \\
Snippet &
  250 &
  5 &
  212 &
  9.9 &
  52.4 &
  53.1 &
  52.8 &
  68.1 &
  67.8 &
  67.9 &
  71.6 &
  81.7 \\
Full Content &
  512 &
  11 &
  208 &
  10.4 &
  42.0 &
  40.7 &
  41.3 &
  58.9 &
  45.4 &
  51.3 &
  63.7 &
  83.2 \\
Full Content &
  256 &
  20 &
  208 &
  10.4 &
  41.3 &
  40.9 &
  41.1 &
  50.4 &
  37.3 &
  42.8 &
  53.9 &
  80.5 \\ \bottomrule
\end{tabular}
\caption{Full results of model performance in different document settings shown in Table~\ref{tab:fulldoc}.}
\label{tab:fulldoc_fullres}
\end{table*}

\section{Experiments on Claim-Split Model}
\label{sec:apdx_claimsplit}
\paragraph{Data for training and evaluation.}
As described in Section~\ref{subsec:claimsplit}, our approach involves fine-tuning mT5 models with ChatGPT outputs. We craft a comprehensive prompt with detailed instructions, as shown in Table~\ref{tab:claimsplit_prompt}. With ChatGPT's feature of structuring \textsc{JSON} outputs, we prompt it to extract all the sub-claims from sentences within the entire summary in a single response, and then split the response into sentence-level claim-split outputs for training and evaluation.
We divide the outputs into training, validation, and test sets aligning with the split of the sample in the WebCiteS dataset. However, since the granularity of sub-claims generated by ChatGPT is not always atomic, we make additional adjustments to the data distribution: we keep all sentences with more than one sub-claim and sample an equal number of sentences with a single sub-claim (either because they are not divisible or because ChatGPT fails to separate their sub-claims). This results in a distribution of 16,158 sentences for training, 2,858 for development, and 1,330 for testing.

\paragraph{Implementation details.}
For fine-tuning, we use the batch size of 64 and the learning rate of 1e-4. We use the AdamW~\citep{loshchilov2018decoupled} optimizer and train the models for 5 epochs. For inference, we use greedy decoding and load the model in half-precision to accelerate evaluation.

\section{Experiments on the AQFS Task.}
\label{sec:apdx_aqfs}

\subsection{Implementation Details}
\label{subsec:apdx_aqfs_implement}
\paragraph{Few-shot prompting (FSP).}
Utilizing the in-context learning abilities of LLMs~\cite{brown2020language}, we construct a prompt with four parts:
\begin{itemize}
\item \textbf{Instruction}: A paragraph that introduces the task and describes specific requirements.
\item \textbf{Demonstration}: An example with the query, source documents, and human-annotated summary as reference.
\item \textbf{Sample to Summarize}: The query and source documents that the model needs to summarize.
\item \textbf{Ending}: An ending statement guiding the model to produce the summary as required.
\end{itemize}

The full prompt is displayed in Table~\ref{tab:fsp_prompt}.

\paragraph{Supervised fine-tuning (SFT).}
we also fine-tune open-source models in our experiments. To save GPU memory, we condense the above prompt as the input text by shortening the instruction and removing the demonstration. The condensed instruction is present in Table~\ref{tab:instruction}.

For mT5 models, we use the batch size of 64, the learning rate of 1e-4, AdamW as the optimizer, and fine-tune them for 5 epochs.
For other open-source LLMs, we use the same batch size and optimizer, while setting the learning rate to 2e-5 and fine-tuning them for 1 epoch, as we find more epochs lead to the rise of validation loss. All open-source LLMs are trained on 8 NVIDIA A100 40G GPUs using Deepspeed ZeRO Stage-3 framework\citep{Rajbhandari2020zero}. We adopt FP16 mixed precision training~\citep{micikevicius2018mixed} for ChatGLM2 and ChatGLM3, and BF16 mixed precision training for Baichuan2 models, based on their default configurations.

\paragraph{Inference.}
For ChatGPT and GPT-4, we use the default parameters of the OpenAI Chat API; for open-source models, we follow \citet{Gao2023EnablingLL} to use Nucleus sampling~\cite{Holtzman2020The} with top\_p=0.95. We load open-source LLMs in either FP16 or BF16 precision to accelerate inference and save GPU memory.

\paragraph{Data for the long context setting.}
In Section~\ref{subsec:long_context}, we adopt a long-context setting where the models are provided with the full content of web pages to summarize. We chunk the web page into documents with a maximum length of 512 or 256 characters. The chunking is performed at the sentence level, where we try to avoid splitting a single sentence into multiple documents. Web pages shorter than the maximum document length are directly taken as the documents.

\begin{table*}[t]
    \includegraphics[width=\linewidth]{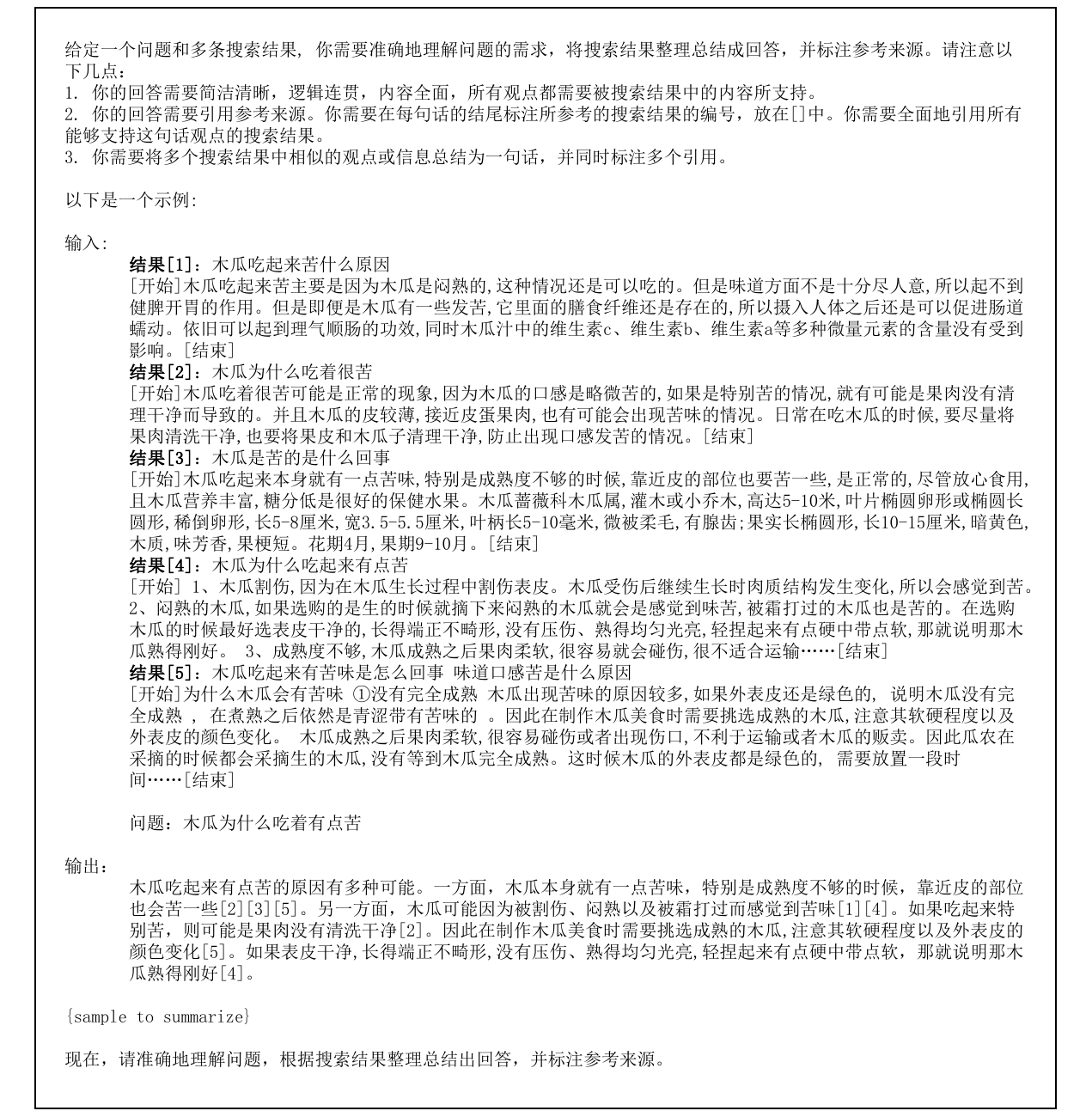}
    \caption{The prompt we use for few-shot prompting with full instruction. Table~\ref{tab:fsp_prompt_en} presents the translation.}
    \label{tab:fsp_prompt}
\end{table*}

\begin{table*}
    \includegraphics[width=\linewidth]{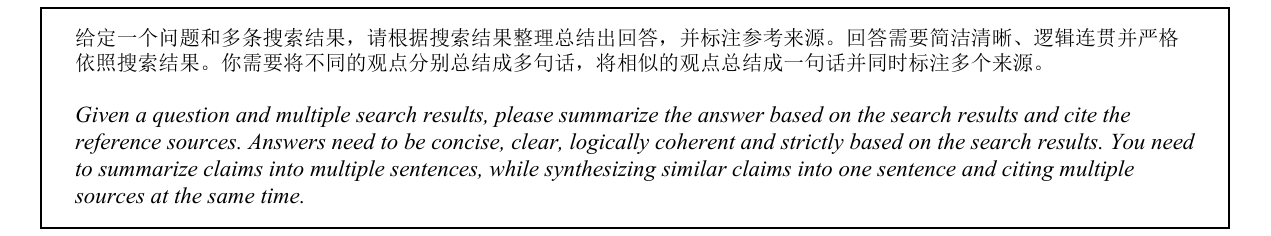}
    \caption{The condensed instruction used for supervised fine-tuning. The translation is in italic text.}
    \label{tab:instruction}
\end{table*}

\begin{table*}
    \includegraphics[width=\linewidth]{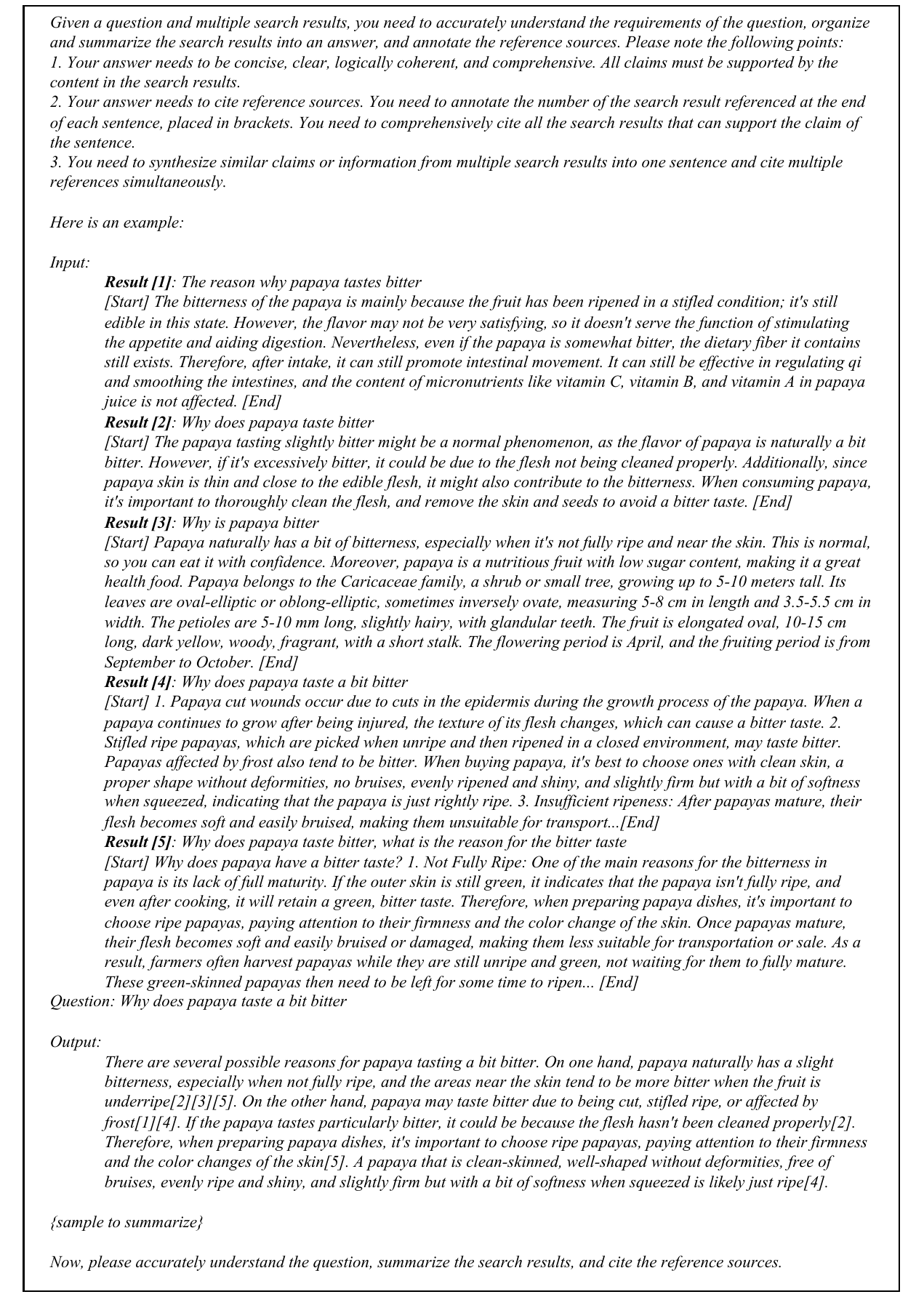}
    \caption{The translation of the prompt in Table~\ref{tab:fsp_prompt}.}
    \label{tab:fsp_prompt_en}
\end{table*}

\begin{table*}
\centering
    \includegraphics[width=\linewidth]{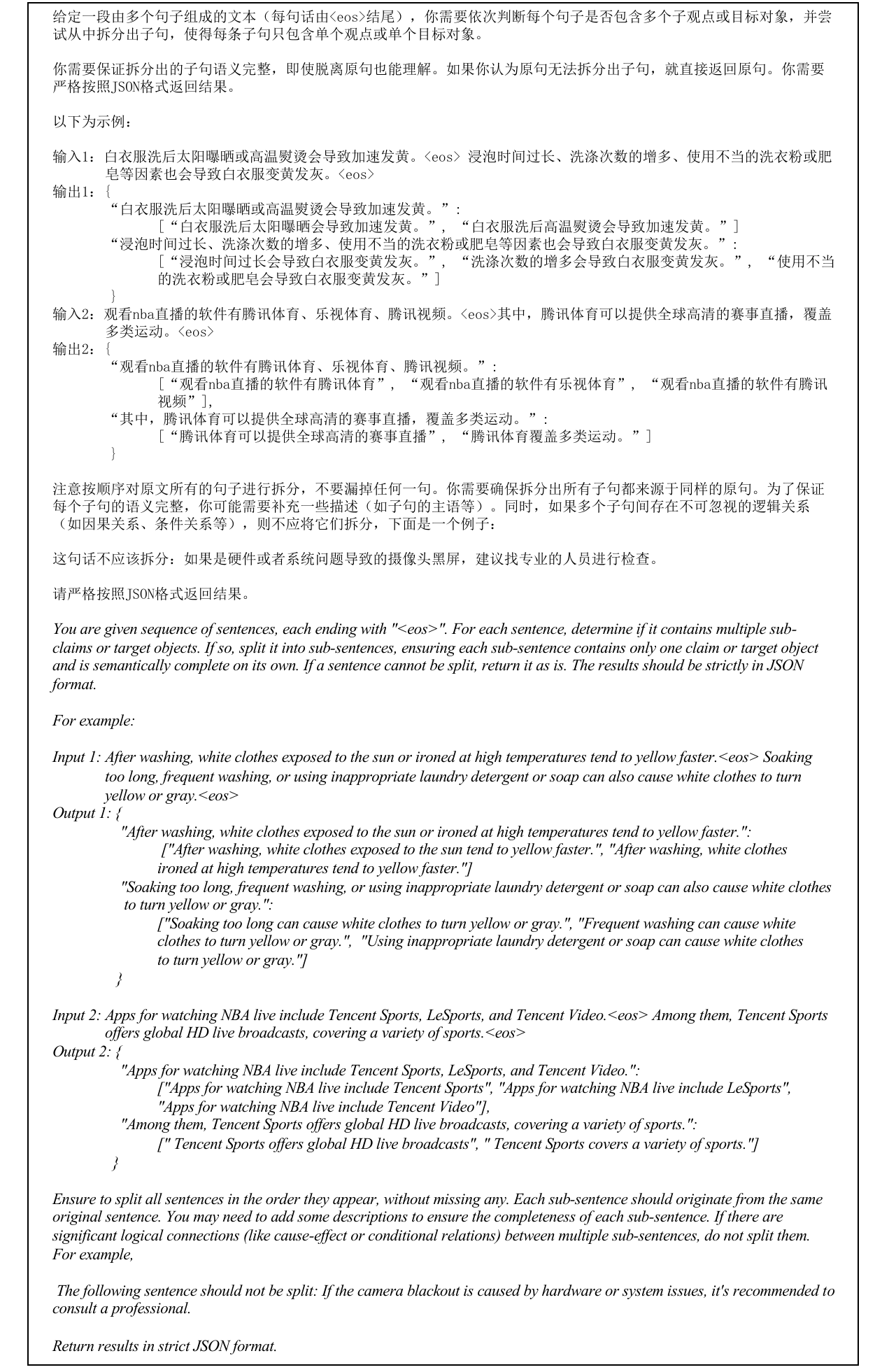}
\caption{The prompt we used to split sub-claims with ChatGPT. The translation is in italic text.}
\label{tab:claimsplit_prompt}
\end{table*}

\label{sec:apdx_prompts}
\end{CJK*}
\end{document}